%% file: main.tex
\pdfoutput=1
\documentclass{article}

\usepackage[final]{corl_2026}
\usepackage{amsmath,amssymb,amsthm,mathtools}
\usepackage{booktabs}
\usepackage{graphicx}
\usepackage{hyperref}
\usepackage{xcolor}
\usepackage{enumitem}
\usepackage{float}
\usepackage{bm}
\usepackage{multirow}
\floatstyle{ruled}
\newfloat{algorithm}{t}{loa}
\floatname{algorithm}{Algorithm}
\usepackage{xcolor}         %
\usepackage{graphicx}
\usepackage{subcaption}
\usepackage{algpseudocode}

\newtheorem{theorem}{Theorem}
\newtheorem{proposition}[theorem]{Proposition}

\theoremstyle{definition}

\newcommand{\R}{\mathbb{R}}
\newcommand{\E}{\mathbb{E}}
\newcommand{\N}{\mathcal{N}}
\newcommand{\C}{\mathcal{C}}

\newcommand{\pth}{\pi_\theta}

\newcommand{\indic}[1]{\mathbf{1}_{#1}}
\newcommand{\proj}{\mathcal{P}}
\newcommand{\AMR}{\texttt{CS-Pi}}

\title{\AMR: Constraint-Satisfying Flow Policy Sampling}

\author{
  Soumojit Bhattacharya \quad Oliver Kroemer$^{\dagger}$ \quad Shubham Tulsiani$^{\dagger}$\\
  Robotics Institute, Carnegie Mellon University\\
  \texttt{\{soumojib, okroemer, stulsian\}@andrew.cmu.edu}\\
  {\small $^{\dagger}$Equal advising}
}
\begin{document}
\maketitle

\begin{abstract}

A flow-matching policy deployed in the field must often satisfy a constraint
that was absent from its demonstrations, such as avoiding an obstacle it has
never seen. Retraining for every new constraint is impractical, so the
constraint must be enforced at inference time. Enforcing it must not discard
what the policy learned: an action the policy would rarely sample may avoid the
obstacle yet fail the task. We therefore want actions that are both
\emph{safe}, satisfying the constraint, and \emph{plausible}, remaining likely
under the policy. Gradient guidance, the most common approach, steers noisy
states, so it must evaluate its cost on the policy's running estimate of the
clean action; when the policy is uncertain between passing left or right of an
obstacle, this estimate is their average and can pass straight through it. We
instead build on rejection and projection. Rejection redraws actions until one
is safe: every output is plausible, but the wait is intractable when safe
actions are rare. Projection moves an action to the nearest safe one: it is
efficient, but the result need not be plausible. Our key insight is that
flow-matching sampling lets us combine the two, keeping projection's efficiency
and rejection's plausibility. \AMR{}, our draw-project-renoise sampler,
projects each draw to the nearest safe action, adds noise back scaled to how
far the projection moved it, and re-runs the unmodified policy until a draw is
safe. Replacing only the projection step extends \AMR{} to continuous costs.
Across a 2D diagnostic, a 7-DoF shelf pick-and-place, an orientation
preference, and a zero-shot Franka deployment, \AMR{} reaches $81\%$ safe
success on the shelf and $55\%$ on hardware, against $55\%$ and $40\%$ for
projection, while staying plausible.

\end{abstract}

\input{intro_st}

\section{Related Work}
\label{sec:related}
\vspace{-6pt}
\paragraph{Gradient guidance.}
In the robotic settings we consider, constraints and preferences are typically
expressed as analytic costs on the action, such as a signed-distance collision
cost, a joint-limit penalty or an orientation error. The standard way to steer
a diffusion or flow-matching policy \citep{lipman2023flowmatchinggenerativemodeling}
toward low cost at deployment is gradient guidance, which nudges each denoising
step along the gradient of that cost: classifier-free guidance
\citep{ho2022classifierfreediffusionguidance,dhariwal2021diffusionmodelsbeatgans},
DPS \citep{chung2024diffusionposteriorsamplinggeneral}, $\Pi$GDM
\citep{song2023pseudoinverse}, MPGD \citep{he2023manifoldpreservingguideddiffusion},
AdjointDPM \citep{pan2024adjointdpmadjointsensitivitymethod}, Diffuser
\citep{janner2022planningdiffusionflexiblebehavior}, and Lan-o3dp
\citep{li2025languageguidedobjectcentricdiffusionpolicy}. But the cost is
meaningful only on a clean action, so each method evaluates it at a proxy, the
policy's running estimate of the clean action. The proxy is accurate when the
cost changes slowly and the policy has settled on one mode, and it breaks down
otherwise, pushing the sample toward implausible actions
(App.~\ref{app:guidance-fail}). Trust Sampling
\citep{huang2024constraineddiffusiontrustsampling} takes several guidance steps
per denoising step, capping their number by how far the proxy can be trusted at
that noise level and stopping early when the sample leaves the region the
policy was trained on. It shares our aim of keeping samples plausible, but every
step still follows the proxy's gradient.

\vspace{-6pt}

\paragraph{Rejection, reweighting, and projection.}
A constraint can also be met without following the proxy's gradient, by acting
on the policy's clean actions. Rejection sampling redraws complete actions until
one is safe: every output is plausible, but the wait is intractable when safe
actions are rare. Sequential Monte Carlo (SMC) samplers reweight and resample a
population of particles rather than discarding them (the Twisted Diffusion
Sampler \citep{wu2024practicalasymptoticallyexactconditional}, Feynman--Kac
samplers \citep{skreta2025feynmankaccorrectorsdiffusionannealing,singhal2025generalframeworkinferencetimescaling});
they need more particles as the safe set shrinks, and the weight that steers
each particle at a noisy step is still evaluated at the proxy
(App.~\ref{app:guidance-fail}). Projection moves a sample to the nearest safe
action without any cost gradient, either at every denoising step
\citep{zhang2025constraineddiffuserssafeplanning} or once, on the final action,
which is then executed; in both cases the projected action need not be
plausible. \AMR{} also avoids the proxy's gradient but combines rejection and
projection: it projects the final action, then renoises the projection rather
than executing it and re-denoises it with the policy, returning a plausible
draw closer to the safe set without rejection's wait or SMC's particle budget.

\vspace{-4pt}

\paragraph{Control-barrier-function samplers.}
SafeDiffuser~\citep{xiao2023safediffusersafeplanningdiffusion} enforces forward
invariance by solving a control-barrier-function (CBF) quadratic program at
each denoising step; SafeFlowMatcher~\citep{yang2026safeflowmatchersafefastplanning}
alternates flow prediction with barrier-based correction; and SAD-Flower
\citep{huang2026sadflowerflowmatchingsafe} augments a flow with a virtual safe
control input and a learned dynamics model. These methods are designed to
certify safety under their respective assumptions, but the certificate says
nothing about plausibility. Their corrections continuously deform the current
sample: when a constraint blocks the mode the sample has committed to, a
correction can push it out of the unsafe region but does not redraw it from
another mode of the policy. \AMR{} instead projects, renoises and re-denoises
with the policy, which allows such a mode switch while keeping each draw
plausible.

CBF samplers also require a differentiable barrier with the structure needed to
evaluate its Lie derivatives and relative degree; SAD-Flower additionally
assumes a learned dynamics model. \AMR{} requires only a projection onto the
safe set, which enables its zero-shot use with existing policies, such as $\pi_{0.5}$, and obstacle bounding
boxes estimated from vision (Sec.~\ref{sec:exp-hardware}). 
We test SafeDiffuser and SafeFlowMatcher on the same policies and report them in
Tables~\ref{tab:d3il-main} and~\ref{tab:shelf} and App.~\ref{app:sphere}.
SAD-Flower is not evaluated as it requires a learned dynamics model, which our
setups do not provide.

\vspace{-6pt}

\paragraph{Deployment-time safety for vision-language-action models.}
Recent work shares our practical motivation of enforcing constraints on a
pretrained vision-language-action (VLA) policy at deployment. OmniGuide
\citep{song2026omniguideuniversalguidancefields} steers flow-matching VLAs
\citep{intelligence2025pi05visionlanguageactionmodelopenworld,nvidia2025gr00tn1openfoundation}
toward collision avoidance and semantic targets with gradient guidance on
differentiable energy fields, and thus evaluates its cost at the proxy.
\textsc{VLSA} \citep{hu2025vlsavisionlanguageactionmodelsplugandplay} instead
acts once, on the final action, much as projection does: a CBF filter changes
the VLA's output action as little as possible to satisfy the barrier condition,
and the robot executes the filtered action. Its authors report that this filter
causes a safety-induced distribution shift. Our hardware experiment uses a
similar setup, a $\pi_{0.5}$ policy on a Franka arm, and there \AMR{} improves
on both gradient guidance and projection (Sec.~\ref{sec:exp-hardware}). 

\vspace{-6pt}

\paragraph{Partial renoising.}
Adding noise to an input and denoising it again is an established way to make
an edited input plausible under a generative model. \textsc{SDEdit}
\citep{DBLP:journals/corr/abs-2108-01073} renoises a user's stroke painting and
denoises it into a realistic image; \textsc{RePaint}
\citep{lugmayr2022repaintinpaintingusingdenoising} renoises around known pixels
to inpaint the rest; and \textsc{DemoDiffusion}
\citep{park2026demodiffusiononeshothumanimitation} renoises a human
demonstration retargeted to the robot into actions the policy would plausibly
produce. In each, the renoised input is given in advance and the amount of
noise is set by hand, once. \AMR{} instead renoises the policy's own draw after
projecting it onto the safe set, sets the amount of noise from how far the
projection moved the draw, and repeats until a draw is safe. This turns a
one-shot edit into a loop that drives the draw toward the safe set
(Sec.~\ref{sec:method}).

\vspace{-8pt}
\section{Approach}
\label{sec:method}
\vspace{-4pt}

We want to sample from a flow-matching policy under a constraint supplied only at deployment time. The action we draw must satisfy two equally important requirements: it must be \emph{safe}, satisfying the constraint, and \emph{plausible}, remaining likely under the policy. An action the policy would essentially never draw, even one that satisfies the constraint, may be jerky, kinematically infeasible, or task-inappropriate.

\textbf{Setup and notation.}
The policy
$\pth$ generates a clean action $\mathbf{a} \in \R^D$ by integrating a
learned velocity field along an ODE, from Gaussian noise
$\mathbf{a}_0 \sim \N(\mathbf{0}, I)$ at time $t = 0$ to the clean
action at $t = 1$. We write $\mathbf{a} = \pth(\mathbf{a}_n;\, n)$ for
integrating from an intermediate state $\mathbf{a}_n$ at time
$t = n \in [0, 1]$ to $t = 1$, so an ordinary draw is
$\mathbf{a} = \pth(\mathbf{a}_0;\, 0)$ with
$\mathbf{a}_0 \sim \N(\mathbf{0}, I)$, abbreviated
$\mathbf{a} \sim \pth$; we overload $\pth(\mathbf{a})$ for the marginal
density it induces over clean actions. At deployment we are given a
safe set $\C \subseteq \R^D$, absent at training, through its
Euclidean projection
$\proj(\mathbf{a}) = \arg\min_{\mathbf{a}' \in \C} \|\mathbf{a}' - \mathbf{a}\|$, which is 
closed-form or a cheap iterative solve for the constraints we
consider. Our goal is to draw $\mathbf{a} \sim \pth$ with
$\mathbf{a} \in \C$. We generalize our method to admit continuous cost functions at the end
of the section.

\textbf{Overview.}
We want samples that are both safe and plausible. Rejection sampling
and projection are two classical approaches, each achieving one of
these: rejection redraws unmodified policy
samples until one is safe, so every draw is plausible, but almost
all are wasted when safe actions are rare; projection maps the clean sample to
the nearest point in the safe set, so every action is safe, but the
projected action is one the policy is unlikely to produce. Our insight is to
strike a balance between the two, projection's efficiency and
rejection's plausibility, by using the projection not as the action
to execute but as an anchor for the next draw. Concretely, we draw a
clean action, project it onto $\C$, renoise the projection to a noise
level set by how far the projection moved the draw (the projection residual), 
and re-denoise from there; we iterate until a draw is itself safe.

\textbf{From projection to a resampling target.}
Projection gives us a useful starting point for the next draw. The
projected sample
$\mathbf{a}_c^{(i)}=\proj(\hat{\mathbf{a}}^{(i)})$ of the current clean
draw $\hat{\mathbf{a}}^{(i)}$, which we call the \emph{anchor}, is cheap to compute and safe by
construction, and it sits closer than $\hat{\mathbf{a}}^{(i)}$ to the
safe actions the policy is likely to draw. It therefore
tells us \emph{where} the next draw should concentrate, i.e. near
$\mathbf{a}_c^{(i)}$, without executing $\mathbf{a}_c^{(i)}$.
The corresponding idealized target biases the policy toward
$\mathbf{a}_c^{(i)}$ while preserving its own structure: 
{\setlength{\abovedisplayskip}{3pt}%
 \setlength{\belowdisplayskip}{3pt}%
 \setlength{\abovedisplayshortskip}{0pt}%
 \setlength{\belowdisplayshortskip}{3pt}%
\begin{equation}
  q^{(i)}(\mathbf{a}) \;\propto\; \pth(\mathbf{a}) \cdot
  \N(\mathbf{a};\,\mathbf{a}_c^{(i)},\,(\sigma^{(i)})^2 I).
  \label{eq:proposal}
\end{equation}}%
Drawing from Eq.~\eqref{eq:proposal}, projecting the result, and recentering
the next target on that projection defines one iteration of a loop
that terminates when a draw lands in $\C$.

\textbf{Setting $\sigma^{(i)}$ from the projection residual.}
We set $\sigma^{(i)}$ from the projection residual
$\delta^{(i)}=\|\hat{\mathbf a}^{(i)}-\mathbf a_c^{(i)}\|$ itself,
because $\delta^{(i)}$ already encodes how much trust to place in
$\mathbf a_c^{(i)}$ as an anchor.
A small $\delta^{(i)}$ means
the current draw was nearly safe, so plausible actions likely lie near
$\mathbf a_c^{(i)}$ and the next draw should
concentrate there. A large $\delta^{(i)}$ means the current draw was far
from $\C$, so $\mathbf a_c^{(i)}$ may sit in a region the policy assigns vanishing
mass; the next draw should be loose enough to give a different policy
mode (e.g., passing right of an obstacle rather than left) a chance to be reached.
We therefore set $\sigma^{(i)}=\delta^{(i)}/k$ for a positive coefficient $k$,
the only hyperparameter introduced in the loop. The hyperparameter $k$
slides between the two extremes: $k\to0$ gives
$\sigma^{(i)}\to\infty$ and recovers rejection, while $k\to\infty$
gives $\sigma^{(i)}\to0$ and recovers projection.

\textbf{Realising the target via partial renoising.}
Sampling from Eq.~\eqref{eq:proposal} for an arbitrary $\sigma^{(i)}$ is
intractable in general as the policy density $\pth(\mathbf a)$ is implicit,
defined only through the ODE, and its product with a Gaussian has no
closed form. We therefore renoise the projected point to the intermediate time
$n^{(i)}=k/(k+\delta^{(i)})$,
{\setlength{\abovedisplayskip}{3pt}%
 \setlength{\belowdisplayskip}{3pt}%
 \setlength{\abovedisplayshortskip}{0pt}%
 \setlength{\belowdisplayshortskip}{3pt}%
\begin{equation}
  \mathbf a_n^{(i)}
  =n^{(i)}\mathbf a_c^{(i)}
   +(1-n^{(i)})\boldsymbol\epsilon^{(i)},\qquad
  \boldsymbol\epsilon^{(i)}\sim\N(\mathbf{0},I),
  \label{eq:renoise}
\end{equation}}%
and re-denoise from $\mathbf a_n^{(i)}$. This choice of $n^{(i)}$ ties
the amount of noise to $\sigma^{(i)}$:
$(1-n^{(i)})/n^{(i)}=\delta^{(i)}/k=\sigma^{(i)}$, so a larger residual
adds more noise. (App.~\ref{app:sampling-law}) states
the exact rule for the next draw under this construction and its scope.

\textbf{Iterating to safety.}
We pass each new draw back as the next $\hat{\mathbf{a}}$: project,
renoise, re-run the policy, repeat, until the draw lands in $\C$. We refer to this
draw-project-renoise loop as \emph{Constraint-Satisfying Flow Policy Sampling}
(\AMR{}; Fig.~\ref{fig:concept}), stated in
Algorithm~\ref{alg:amr}; it terminates when $\delta \leq \tau$, i.e., once a draw lies within $\tau$ of $\C$, or after
a cap of $K_{\max}$ iterations. On average each successive draw lands closer to the safe set (App.~\ref{app:nproj}), so the residual $\delta$ shrinks (resulting in a larger $n$) with each iteration, bringing the current draw closer to a safe and plausible draw.

\begin{algorithm}[htp]
\caption{Constraint-Satisfying Flow Policy Sampling}
\label{alg:amr}
\begin{algorithmic}[1]
\State Draw $\mathbf a_0 \sim \N(\mathbf 0, I)$;\;
       $\hat{\mathbf a}^{(0)} \gets \pth(\mathbf a_0;\, 0)$
       \Comment{initial policy draw}
\For{$i = 0, \dots, K_{\max}-1$}
    \State $\mathbf a_c^{(i)} \gets \proj(\hat{\mathbf a}^{(i)})$;\;
           $\delta^{(i)} \gets \|\hat{\mathbf a}^{(i)} - \mathbf a_c^{(i)}\|$
    \If{$\delta^{(i)} \leq \tau$}
        \State \Return $\hat{\mathbf a}^{(i)}$
    \EndIf
    \State $n^{(i)} \gets k / (k + \delta^{(i)})$;\;
           $\boldsymbol\epsilon^{(i)} \sim \N(\mathbf 0, I)$
    \State $\mathbf a_n^{(i)} \gets n^{(i)} \mathbf a_c^{(i)}
           + (1 - n^{(i)}) \boldsymbol\epsilon^{(i)}$
    \State $\hat{\mathbf a}^{(i+1)} \gets \pth(\mathbf a_n^{(i)};\, n^{(i)})$
           \Comment{re-denoise from time $n$}
\EndFor
\State \Return $\proj(\hat{\mathbf a}^{(K_{\max})})$
       \Comment{optional terminal projection}
\end{algorithmic}
\end{algorithm}

\vspace{-6pt}

\textbf{Incorporating soft preferences as continuous costs.}
So far we have enforced a \emph{hard} constraint, i.e. an action either lies
in $\C$ or it does not. Deployment also brings softer \emph{preferences}, behaviors to
favor rather than boundaries to clear, such as keeping the gripper
near-vertical while carrying a full cup (Sec.\ref{sec:exp-mpd9}), and we generalize the same
loop to express these preferences alongside hard constraints. This generalization
needs only one change: \AMR{} uses the projection only to produce an anchor for
the next draw; for a continuous cost $L$, that anchor comes instead
from the proximal operator
\begin{equation}
  \mathrm{prox}_{\beta L}(\hat{\mathbf{a}})
  \;=\; \arg\min_{\mathbf{a}' \in \R^D}\,\Big[\, L(\mathbf{a}') + \tfrac{1}{2\beta}\,\|\mathbf{a}' - \hat{\mathbf{a}}\|^2 \,\Big],
  \label{eq:prox}
\end{equation}
which trades reducing $L$ against staying close to the current draw
$\hat{\mathbf{a}}$, with $\beta$ setting the balance; the rest of the
algorithm is unchanged. Hard constraints and continuous costs are then composed by being applied in sequence to the same clean draw, with the operators encoding priority by their order. For example, \AMR{} can first pull toward the preferred orientation with the proximal operator and then project onto the safe set, so the constraint takes precedence. A gradient method would instead have to merge both into a single objective.

\textbf{Optional terminal projection.} If no draw lies within $\tau$ of $\C$
after $K_{\max}$ iterations, \AMR{} can project the last draw and execute the
result, trading plausibility for safety when safety must come first. Even then
the loop pays off: in simulation, the executed actions are more likely under
the policy than a typical one-shot projection.
App.~\ref{app:executed-guarantee} studies how often the terminal projection
is triggered and how likely the resulting actions are.

\vspace{-8pt}
\section{Experiments}
\label{sec:experiments}
\vspace{-8pt}
Throughout, we measure two things for each sampler: whether it satisfies
the constraint and whether its samples are plausible. We
evaluate four settings: a 2D navigation task where the whole action
distribution is easy to visualize, a 7-DoF shelf pick-and-place, an
orientation-preference cost (\textsc{3D-Orient}), and a
real-world Franka deployment; App.~\ref{app:sphere} adds a 3D
obstacle-avoidance benchmark, \textsc{Sphere-Avoid}.

Our primary metric is the \emph{safe success rate} (Safe-SR): the fraction of
rollouts that satisfy the constraint \emph{and} complete the task.
Plausibility is measured by logL, the mean log-likelihood $\pth$ assigns to a
method's actions (nats/dim; higher is more plausible).
Compute is reported as \textsc{NFE}, the flow-network forward passes per
rollout (not comparable across benchmarks). We compare against gradient
guidance, Trust Sampling (step size $w$, at most $J$ guidance steps per
denoising step), rejection sampling (redraw budget $B$), projection, the
unconstrained policy, and two CBF samplers (SafeDiffuser~\citep{xiao2023safediffusersafeplanningdiffusion} and
SafeFlowMatcher~\citep{yang2026safeflowmatchersafefastplanning}), each reported at the best value of its main
hyperparameters (App.~\ref{app:hpsweep}). \AMR{} has a single hyperparameter,
$k$, and is robust across a broad range of it (App.~\ref{app:ktune}). It uses
the terminal projection on every obstacle benchmark but not on
\textsc{3D-Orient}, where the target is a preference.

\vspace{-3pt}
\subsection{2D \textsc{Avoidance}: constraint satisfaction without distribution shift} 
\vspace{-4pt}
\label{sec:exp-2d-main}

\begin{figure}[htp]
\centering
\includegraphics[width=0.8\linewidth]{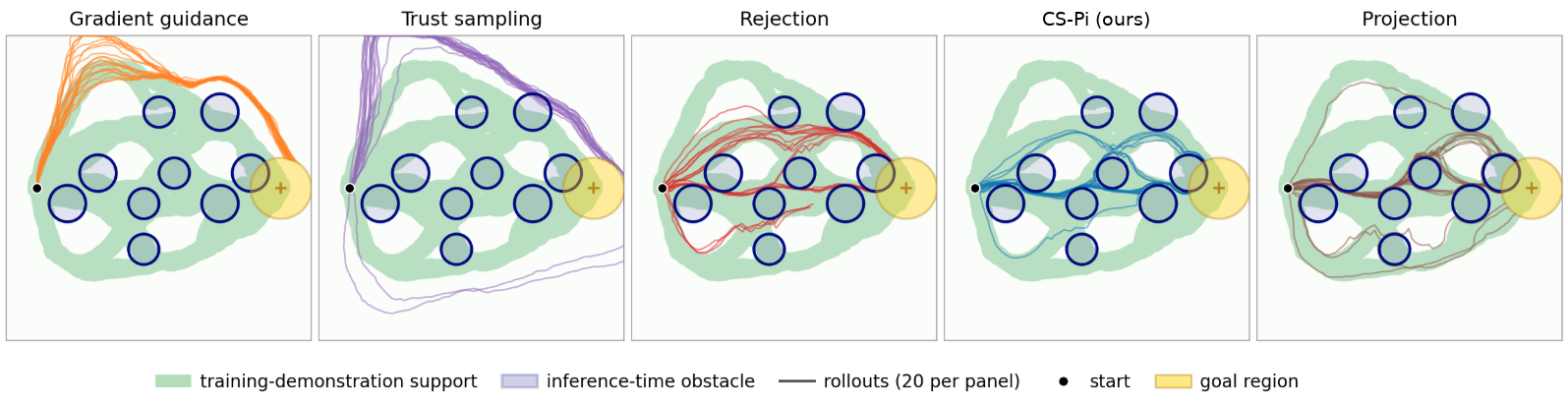}
\caption{\textbf{2D \textsc{Avoidance} trajectory distributions, one
scenario per method.} $20$ rollouts overlaid on the
demonstration support (shaded); the obstacles are outlined.
Gradient guidance and Trust Sampling visibly leave the support to reach the goal or
cut through the obstacle; rejection stays on-support but only a few
draws survive; \AMR{} keeps every trajectory
in-distribution while routing around the obstacle. The other five
scenarios are in App.~\ref{app:extra-2d}.}
\label{fig:2d-main}
\end{figure}

\textsc{Avoidance} is a 2D planar environment in which an agent travels
from a fixed start position to a goal region. We collect demonstrations
that reach the goal along \emph{distinct} paths, so the policy is
 multimodal, mirroring real manipulation data where
demonstrations of one task take different routes or have different styles. Because the action space is 2D, we can plot each sampler's full distribution. Plausibility is measured by how many of the
rollouts stay inside the demonstration support: the part of the plane the demonstrations pass through.
At deployment we impose obstacles the policy never saw, in six scenarios
of varying difficulty. Fig.~\ref{fig:2d-main} shows
rollouts for one scenario; the remaining five are in
App.~\ref{app:extra-2d}.

Three methods reach at least $99\%$ safe success: SafeDiffuser, projection and
\AMR{} (Table~\ref{tab:d3il-main}). Among them, \AMR{} keeps the most
rollouts fully on-support ($0.857$) and has the highest log-likelihood
($1.53$), while projection's log-likelihood collapses to $-85.89$. Gradient
guidance, Trust Sampling and rejection stay plausible but reach at most
$79.4\%$ safe success.

\begin{table}[tbp]
\centering\small
\begin{tabular}{lcccr}
\toprule
Method & Safe-SR (\%)\,$\uparrow$ & fully-in-supp.\,$\uparrow$ & logL\,$\uparrow$ & NFE\,$\downarrow$ \\ \midrule
Unconstrained      & $5.2$            & $0.800$          & $\mathbf{1.66}$  & $456$ \\
Gradient guidance  & $67.2$           & $0.410$          & $1.49$           & $298$ \\
Trust Sampling     & $34.9$           & $0.350$          & $1.43$           & $4421$ \\
Rejection          & $79.4$           & $0.750$          & $\mathbf{1.66}$  & $73921$ \\
Projection         & $99.8$           & $0.640$          & $-85.89$         & $\mathbf{52}$ \\
SafeDiffuser       & $\mathbf{100.0}$ & $0.830$          & $1.44$           & $92$ \\
SafeFlowMatcher    & $45.2$           & $0.000$          & $-463$           & $220$ \\
\AMR{}             & $99.0$           & $\mathbf{0.857}$ & $1.53$           & $283$ \\
\bottomrule
\end{tabular}
\caption{\textbf{2D \textsc{Avoidance}, macro over six scenarios}
($200$ rollouts per cell; best hyperparameter per method).
\emph{Safe-SR}: rollouts reaching the goal collision-free.
\emph{fully-in-supp.}: fraction of rollouts whose \emph{every} waypoint
lies inside the demonstration support.
\emph{logL}: mean log-likelihood under $\pth$.}
\label{tab:d3il-main}
\end{table}

\paragraph{Composing a hard constraint with a soft preference.}
A safe set cannot express a \emph{preference} among feasible
actions. On the same 2D \textsc{Avoidance} environment we add to the
obstacle constraint a soft preference toward a height
$y_{\mathrm{ref}}{=}0.7$. \AMR{} applies the two in sequence by priority,
whereas the baselines must fold them into one weighted cost; only \AMR{}
satisfies both (App.~\ref{app:extra-2d}, Fig.~\ref{fig:softhard-2d}).

\vspace{-2pt}
\subsection{Shelf pick-and-place with deployment-time obstacles}
\label{sec:exp-shelf}
\vspace{-4pt}
The shelf benchmark tests whether a sampler can satisfy a
deployment-time obstacle constraint \emph{without} discarding the task
semantics encoded by the policy. A 7-DoF Franka picks a 4\,cm cube from a three-rung shelf and places it
on the table; at inference we add obstacles absent at training
(App.~\ref{app:shelf}). A rollout is \texttt{safe and successful} if it is
collision-free, picks the cube, and places it. \AMR{} and projection
share the same L-BFGS \citep{Liu1989OnTL} proximal step, so any gap between them is due to the iterative loop of \AMR{}.

\AMR{} reaches the highest safe success rate ($81.3\%$;
Table~\ref{tab:shelf}) and is the only method whose actions score above the
unconstrained policy in mean log-likelihood; projection and the gradient
baselines satisfy the obstacle only by making the chunk implausible. \AMR{},
projection, gradient guidance and SafeFlowMatcher collide on at most $8\%$ of
rollouts, so among them performance depends mainly on whether the gripper still
reaches the cube. Projection and the gradient baselines avoid the obstacles by
moving the arm away from them, and often away from the cube as well; \AMR{}
instead re-denoises a trajectory that passes between the obstacles and still
reaches the cube ($84.7\%$ goal reaching, above even the unconstrained policy's
$75.3\%$).

\begin{table}[hbtp]
\centering\small
\setlength{\tabcolsep}{4.5pt}
\begin{tabular}{lccccr}
\toprule
Method & Safe-SR (\%)\,$\uparrow$ & Goal reaching (\%)\,$\uparrow$ & Collision (\%)\,$\downarrow$ & logL\,$\uparrow$ & NFE\,$\downarrow$ \\
\midrule
Unconstrained                          & $3.3$           & $75.3$          & $80.0$          & $15.02$          & $240$  \\
Gradient guidance ($\alpha{=}0.20$)    & $32.0$          & $32.0$          & $\mathbf{0.7}$  & $10.42$          & $240$  \\
Trust Sampling ($w{=}0.1$, $J{=}5$)    & $53.3$          & $60.7$          & $12.0$          & $12.23$          & $1104$ \\
Rejection ($B{=}100$)                  & $34.7$          & $68.0$          & $38.7$          & $14.22$          & $8745$ \\
Projection                             & $55.3$          & $58.7$          & $6.0$           & $11.41$          & $240$  \\
SafeDiffuser                           & $13.3$          & $16.7$          & $19.3$          & $7.50^\dagger$   & $240$  \\
SafeFlowMatcher                        & $44.7$          & $48.7$          & $8.0$           & $10.90$          & $432$  \\
\AMR{} ($k{=}50$, $K_{\max}{=}5$)      & $\mathbf{81.3}$ & $\mathbf{84.7}$ & $3.3$           & $\mathbf{17.96}$ & $271$  \\
\bottomrule
\end{tabular}
\caption{\textbf{Shelf pick-and-place, $5$ scenarios, $150$ tasks
($30$/scenario).} \emph{Safe-SR}: collision-free rollouts that pick and place
the cube. \emph{logL}: mean log-likelihood under $\pth$ (nats/dim; the
unconstrained policy scores $15.02$). \textsc{NFE}: flow-model calls per
trajectory. $^\dagger$Median: a few diverged SafeDiffuser chunks make its mean
uninformative.}
\label{tab:shelf}
\end{table}
\vspace{-4pt}

\vspace{-4pt}
\subsection{3D-Orient: reaching with an orientation preference}
\label{sec:exp-mpd9}
\vspace{-4pt}
\textsc{3D-Orient} tests \AMR{} on a continuous cost. The task is a 7-DoF
reach to a goal sampled anywhere in the robot's reachable workspace; on
top of it we add a soft preference to hold a top-down orientation $R^*$,
encoded as the cost
$\mathcal{L}(q)=\lambda\sum_{t=1}^{H}\lVert R(q_t)-R^*\rVert_F^2$. The test is whether a method can achieve the
orientation without giving up the reach. Projection, rejection and the CBF
samplers require a safe set and do not apply to a continuous cost.

\AMR{} is the only method that does both (Table~\ref{tab:mpd9}):
$65.6\%$ success, with goal reaching ($98.9\%$) and end-effector step
($0.07$\,m) matching the unconstrained policy. Gradient guidance loses the
reach and roughens the path ($<7\%$ success); Trust Sampling keeps the reach
and a high log-likelihood only by barely changing the orientation ($1.1\%$
success). \AMR{}'s lower log-likelihood ($10.21$ vs.\ $13.46$) is the cost of
moving the gripper away from the policy's preferred orientations. Guidance fails because of IK
redundancy: many configurations reach the same point at different
orientations, so the proxy averages them into an invalid configuration with an
uninformative gradient (App.~\ref{app:guidance-fail}).

\begin{table}[htbp]
\centering\small
\begin{tabular}{lccccc}
\toprule
Method (best cfg) & Success\,$\uparrow$ & Goal reaching\,$\uparrow$ & logL\,$\uparrow$ & Max EE step (m) & NFE\,$\downarrow$ \\
\midrule
Unconstrained                         & $2.2\%$           & $\mathbf{98.9\%}$ & $13.46$ & $\mathbf{0.07}$ & $\mathbf{138}$ \\
Gradient guidance ($\alpha{=}0.003$)  & $2.2\%$           & $70.0\%$          & $12.30$ & $0.08$          & $154$ \\
Gradient guidance ($\alpha{=}0.03$)   & $6.7\%$           & $7.8\%$           & $9.62$  & $0.40$          & $160$ \\
Trust Sampling ($w{=}0.001$)          & $1.1\%$           & $\mathbf{98.9\%}$ & $13.44$ & $\mathbf{0.07}$ & $1379$ \\
\AMR{} + EE-prox ($k{=}10$, $K_{\max}{=}3$) & $\mathbf{65.6\%}$ & $\mathbf{98.9\%}$ & $10.21$ & $\mathbf{0.07}$ & $210$ \\
\bottomrule
\end{tabular}
\caption{\textbf{\textsc{3D-Orient}, $90$ rollouts per cell.}
\emph{Success}: rollouts with a step simultaneously within $10$\,cm of
the target position and within $0.36$\,rad of the target orientation.
\emph{Goal reaching}: rollouts ending within $10$\,cm of the target
position.
\emph{logL}: mean log-likelihood under $\pth$; comparisons are meaningful only
at comparable success, since the guidance configurations attain only $2.2\%$
and $6.7\%$ success.
\emph{Max EE step}: largest single-step end-effector displacement.
\emph{EE-prox}: proximal step on the orientation cost (Eq.~\eqref{eq:prox}).
\emph{NFE}: mean number of model evaluations per rollout.}
\label{tab:mpd9}
\end{table}
\vspace{-4pt}

\vspace{-12pt}
\vspace{-4pt}
\subsection{Real-world deployment}
\label{sec:exp-hardware}
\begin{figure}[thbp]
\centering
\includegraphics[width=\linewidth]{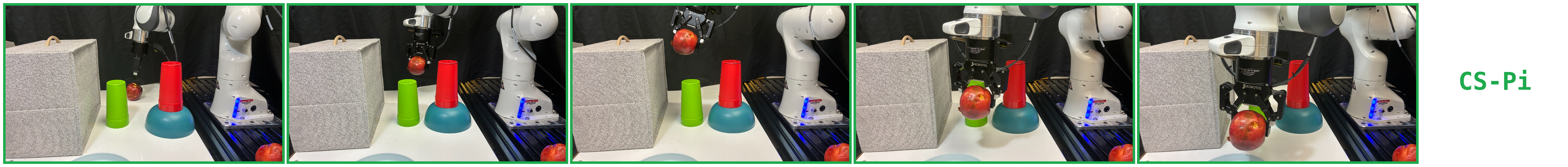}\\[3pt]
\includegraphics[width=\linewidth]{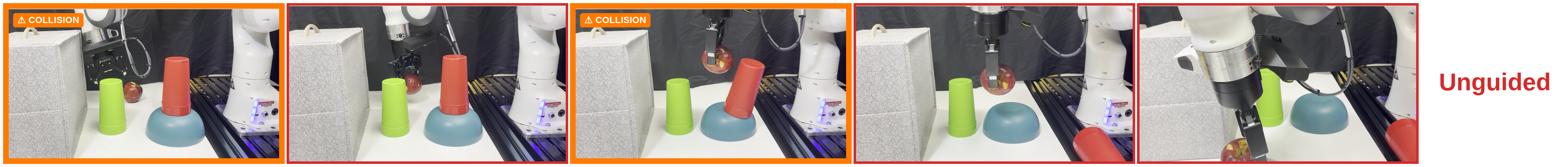}\\[8pt]
\includegraphics[width=\linewidth]{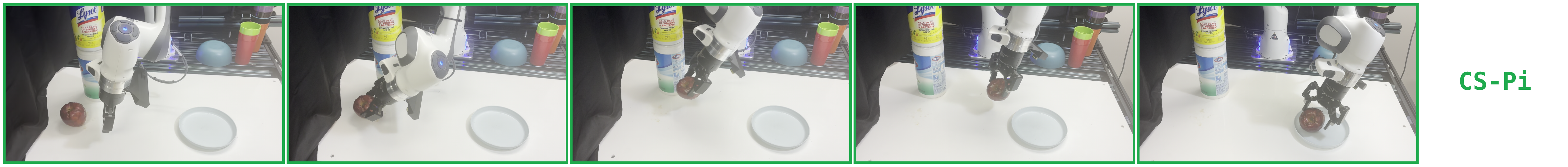}\\[3pt]
\includegraphics[width=\linewidth]{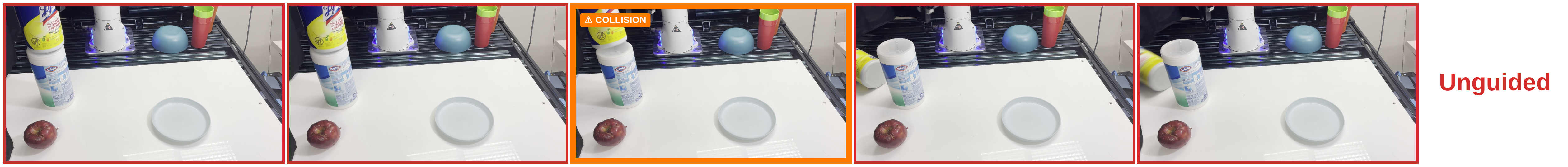}
\caption{\textbf{Hardware rollouts; frames run left-to-right within each
strip.} Green-bordered strips are \AMR{}, which enforces the
deployment-time obstacle constraint and completes the pick-and-place;
red-bordered strips are the same $\pi_{0.5}$ policy run unconstrained. \textbf{Layout~1} (first two): \AMR{} grasps the
apple and places it, while the unconstrained policy clips the cup and loses
the grasp. \textbf{Layout~2} (last two): \AMR{} reaches down and lifts
the apple cleanly, while the unconstrained policy collides with the obstacle.}
\label{fig:hw-rollout}
\end{figure}

\vspace{-4pt}
We evaluate \AMR{} on a Franka arm with a Robotiq 2F-85 gripper, ZED 2
external and ZED Mini wrist cameras, and a DROID-finetuned $\pi_{0.5}$\citep{khazatsky2025droidlargescaleinthewildrobot}
policy used without further finetuning. The task is \emph{pick up the
apple and place it on the blue plate}. A vision pipeline conditioned on
the ZED 2 RGB and depth images generate obstacle bounding boxes that form
the constraint $\C$; deployment is zero-shot. We test over two layouts (Fig.~\ref{fig:hw-rollout}), $10$ trials per method per layout, and
compare \AMR{}, gradient guidance, and projection against the unguided
policy. Table~\ref{tab:hw} reports the safe-success rate, per-step
latency, and the log-likelihood the frozen prior assigns to the
executed actions.

\begin{table}[tbp]
\centering\small
\begin{tabular}{lccccc}
\toprule
& \multicolumn{3}{c}{Safe-SR (\%)\,$\uparrow$} & & \\
\cmidrule(lr){2-4}
Method & Layout 1 & Layout 2 & Averaged & Time/step (s)\,$\downarrow$ & logL\,$\uparrow$ \\
\midrule
Unconstrained     & $20$          & $0$           & $10$          & $0.069$ & $-470.2{\pm}1.0$ \\
Gradient guidance & $40$          & $10$          & $25$          & $0.139$ & $-464.9{\pm}1.4$ \\
Trust Sampling    & $20$          & $20$          & $20$          & $0.135$ & $-474.8{\pm}2.1$ \\
Projection        & $40$          & $40$          & $40$          & $0.099$ & $-473.2{\pm}2.3$ \\
\AMR{}            & $\mathbf{60}$ & $\mathbf{50}$ & $\mathbf{55}$ & $0.143$ & $-471.8{\pm}1.7$ \\
\bottomrule
\end{tabular}
\caption{\textbf{Hardware: \emph{pick the apple, place it on the blue
plate} on a Franka FR3 with $\pi_{0.5}$.} Zero-shot; $\C$ is obstacle
bounding boxes from a ZED 2 RGB-D vision pipeline. Two layouts,
$10$ trials per method per layout ($20$ in total). \AMR{} has the highest
safe success rate in each layout and averaged over both. \emph{logL} is the
log-likelihood $\pth$ assigns to executed actions (the unconstrained policy's
own draws score $-470.2$; higher is more plausible): \AMR{} stays close to the
unconstrained policy, while projection and Trust Sampling fall below it.}
\label{tab:hw}
\end{table}

\AMR{} reaches the highest safe success rate in both layouts ($60\%$
and $50\%$) and averaged over both ($55\%$), versus $40\%$ for
projection, $25\%$ for gradient guidance, $20\%$ for Trust Sampling, and
$10\%$ for the unconstrained policy. Its per-step latency ($0.143$\,s) matches
gradient guidance ($0.139$\,s) and Trust Sampling ($0.135$\,s); projection is
cheaper ($0.099$\,s) but enforces $\C$ with a one-shot projection that drops
its log-likelihood below the unconstrained policy's ($-473.2$ versus
$-470.2$), whereas \AMR{} stays close to it ($-471.8$) while satisfying the
constraint. The runs are small-sample ($10$ trials per layout), but \AMR{}
leads in both layouts.

\vspace{-8pt}

\section{Conclusion}
\label{sec:conclusion}
\vspace{-8pt}
\AMR{} samples safe and plausible actions from a flow-matching policy with a
draw-project-renoise loop: it projects each draw onto the safe set, adds noise
back scaled to how far the projection moved the draw, and re-runs the
unmodified policy. It needs no constraint gradient at noisy states and no
auxiliary network. \AMR{} also extends to continuous
costs by replacing only the projection step with a proximal operator. Across a
2D diagnostic, a 7-DoF shelf pick-and-place, an orientation preference, and a
zero-shot Franka deployment, \AMR{} is the only sampler that satisfies the
constraint while staying plausible wherever the constraint permits. On hardware
it reaches $55\%$ safe success against $40\%$ for projection and $25\%$ for
gradient guidance. A policy can thus meet a safety constraint it never trained
on, without surrendering the behaviour that made it useful.
\vspace{-12pt}
\section*{Limitations}
\label{sec:limitations}
\vspace{-8pt}
Our theory covers a single iteration, we characterize the \emph{per-iteration
sampling distribution}, the distribution of the next draw given the current one
(App.~\ref{app:sampling-law}), but we do not prove that the loop converges. In
practice, \AMR{} does not always reach an action that is both safe and
plausible. Therefore, when no draw lies within $\tau$ of $\C$ after $K_{\max}$ iterations,
it must choose between the last draw, which may be unsafe, and the terminal
projection, which is safe but often less likely under the policy
(App.~\ref{app:executed-guarantee}).

\AMR{} also needs a usable projection $\proj$ (or proximal operator
$\mathrm{prox}_{\beta L}$): the obstacle and joint-limit sets we use admit
closed-form or cheap iterative solutions, but more complex sets need more
complex solvers. It is also only as good as $\C$: on hardware we estimate $\C$
from vision once per rollout and evaluate in static scenes, so dynamic obstacles
and errors from segmentation, depth and registration can misplace or mis-size
$\C$. Finally, like every sampler we evaluate, \AMR{} operates on \emph{action
chunks}, the fixed-length sequences of actions a policy predicts at each query,
and is therefore \emph{myopic}: when the policy commits to a mode that is safe
for the current chunk but has no safe continuation, \AMR{} cannot recover;
trajectory-level planning is out of scope.

\bibliographystyle{corlabbrvnat}
\bibliography{references}

\newpage
\appendix

\section{Project page}\label{app:website}
Qualitative videos of the 2D, simulated, and hardware rollouts, along with
additional visualizations, are available at the project page:
\begin{center}
  \url{https://babayaga840.github.io/CS-pi_site/}
\end{center}

\section{3D \textsc{Sphere-Avoid}: pure obstacle avoidance}\label{app:sphere}
This benchmark complements the shelf result (Sec.~\ref{sec:exp-shelf}) by
isolating obstacle avoidance, with no task-semantic tension between the
constraint and the policy. \textsc{Sphere-Avoid} uses a simulated 7-DoF Franka
Panda and a joint-space action: a flow-matching policy moves between a start and
a goal configuration without colliding with sphere obstacles. The safe set $\C$
is the collision-free set of the swept-sphere geometry plus joint limits. We
test two layouts: L1 ($10$ spheres) and L2 ($16$ spheres). The policy was
trained on demonstrations from both layouts.

\begin{figure}[H]
  \centering
  \begin{subfigure}{0.40\textwidth}
    \centering
    \includegraphics[width=\linewidth]{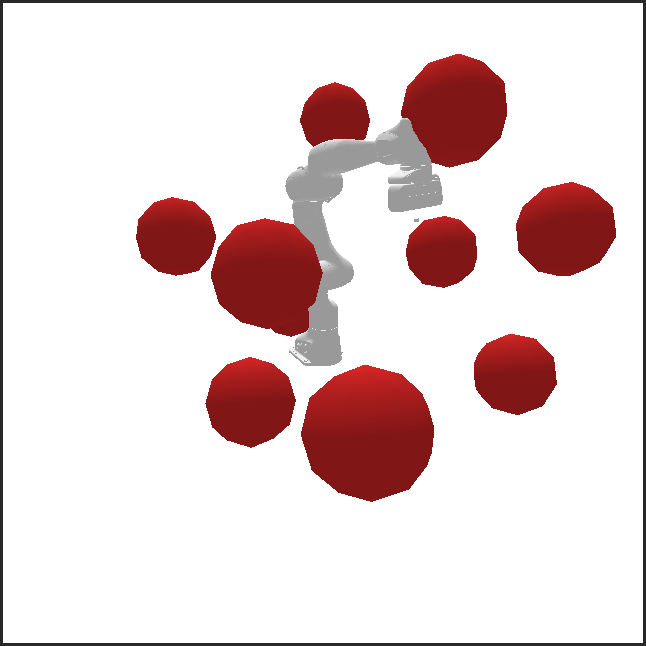}
    \caption{L1 ($10$ spheres)}
    \label{fig:sphere-avoid-level1}
  \end{subfigure}
  \hfill
  \begin{subfigure}{0.40\textwidth}
    \centering
    \includegraphics[width=\linewidth]{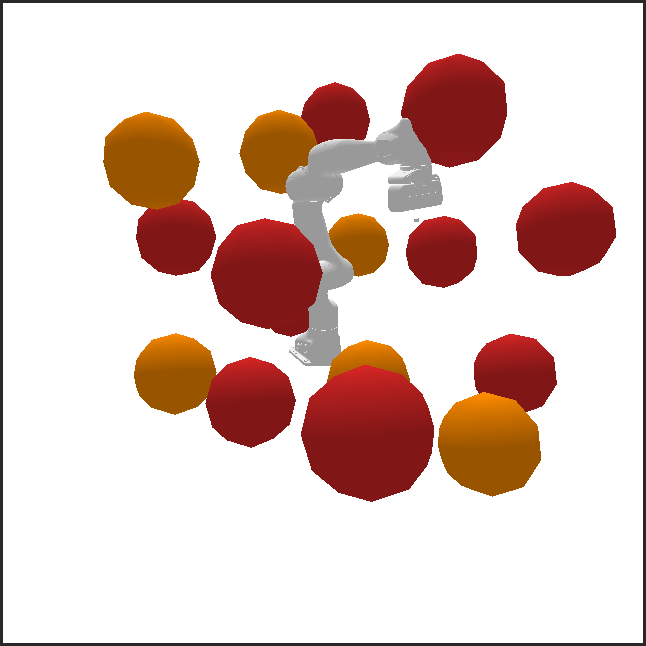}
    \caption{L2 ($16$ spheres)}
    \label{fig:sphere-avoid-level2}
  \end{subfigure}
  \caption{The two \textsc{Sphere-Avoid} layouts (L2 adds six obstacles).}
  \label{fig:spheres-envs}
\end{figure}

\AMR{} with the terminal projection reaches $99.9\%$ safe success on L1 and
$91.8\%$ on L2, the highest in both layouts, and of the methods that enforce the
constraint it loses the least from L1 to L2 ($8.1$ points, against $26.1$ for
projection; Table~\ref{tab:spheres}). The strongest baselines are projection
($88.2\%$ / $62.1\%$ on L1 / L2) and Trust Sampling ($77.0\%$ / $38.7\%$ at
$64$--$69$ NFE), followed by SafeFlowMatcher ($61.8\%$ / $21.9\%$), rejection ($56.7\%$ / $14.1\%$ at
$388$--$492$ NFE), SafeDiffuser ($36.4\%$ / $7.4\%$), gradient guidance
($23.6\%$ / $4.7\%$) and the unconstrained policy ($8.9\%$ / $0.8\%$). \AMR{}
uses $20$--$26$ NFE. Its log-likelihood matches the unconstrained policy on L1
($2.87$ vs.\ $2.88$) and is $0.30$ lower on L2; rejection has the highest
log-likelihood on both layouts.

\begin{table}[H]
\centering\small
\begin{tabular}{lcccccc}
\toprule
& \multicolumn{3}{c}{L1 ($10$ spheres)} & \multicolumn{3}{c}{L2 ($16$ spheres)} \\
\cmidrule(lr){2-4}\cmidrule(lr){5-7}
Method & Safe-SR & logL & NFE & Safe-SR & logL & NFE \\
\midrule
Unconstrained                          & $8.9$           & $2.88$         & $10$  & $0.8$           & $2.84$         & $10$  \\
Gradient guidance ($\alpha{=}0.01$)    & $23.6$          & $2.73$         & $10$  & $4.7$           & $2.32$         & $10$  \\
Trust Sampling ($w{=}0.3$, $J{=}5$)    & $77.0$          & $3.03$         & $64$  & $38.7$          & $2.68$         & $69$  \\
Rejection                              & $56.7$          & $\mathbf{3.47}$& $388$ & $14.1$          & $\mathbf{3.19}$& $492$ \\
Projection                             & $88.2$          & $2.67$         & $10$  & $62.1$          & $2.28$         & $10$  \\
SafeDiffuser                           & $36.4$          & $2.51^\dagger$ & $10$  & $7.4$           & $1.83^\dagger$ & $10$  \\
SafeFlowMatcher                        & $61.8$          & $2.30$         & $12$  & $21.9$          & $1.82$         & $12$  \\
\AMR{} ($k{=}50$, $K_{\max}{=}20$)     & $\mathbf{99.9}$ & $2.87$         & $20$  & $\mathbf{91.8}$ & $2.54$         & $26$  \\
\bottomrule
\end{tabular}
\caption{\textbf{3D \textsc{Sphere-Avoid}.} $120$ tasks, of which $104$ (L1)
and $32$ (L2) have a safe goal configuration; $8$ samples per task.
\emph{logL}: mean log-likelihood under $\pth$ (nats/dim; the unconstrained
policy scores $2.88$ / $2.84$). $^\dagger$Median: a few diverged SafeDiffuser
samples make the mean uninformative ($-119.5$ / $-17.3$). Baselines are at their
best swept configuration (App.~\ref{app:hpsweep}).}
\label{tab:spheres}
\end{table}

\section{Shelf task scenarios}\label{app:shelf}

\begin{figure*}[htp]
  \centering
  \begin{subfigure}{0.19\textwidth}
    \centering
    \includegraphics[width=\linewidth]{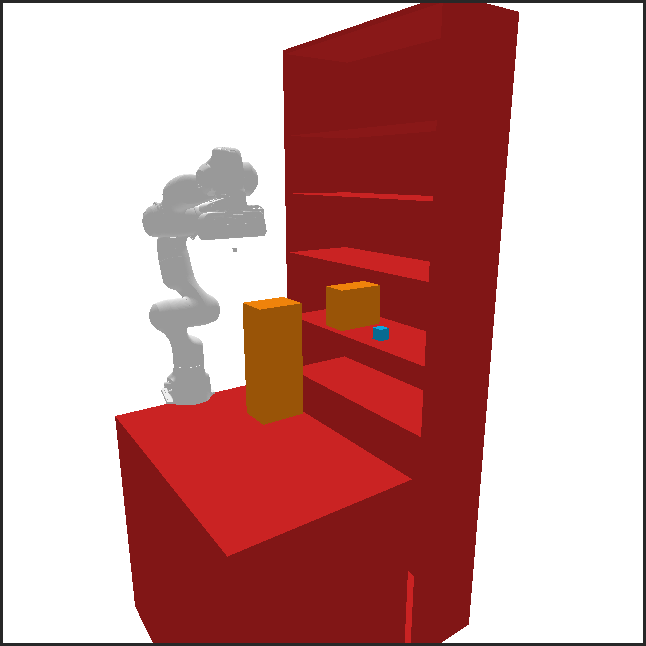}
    \caption{Clutter combo}
    \label{fig:shelf-clutter-combo}
  \end{subfigure}
  \hfill
  \begin{subfigure}{0.19\textwidth}
    \centering
    \includegraphics[width=\linewidth]{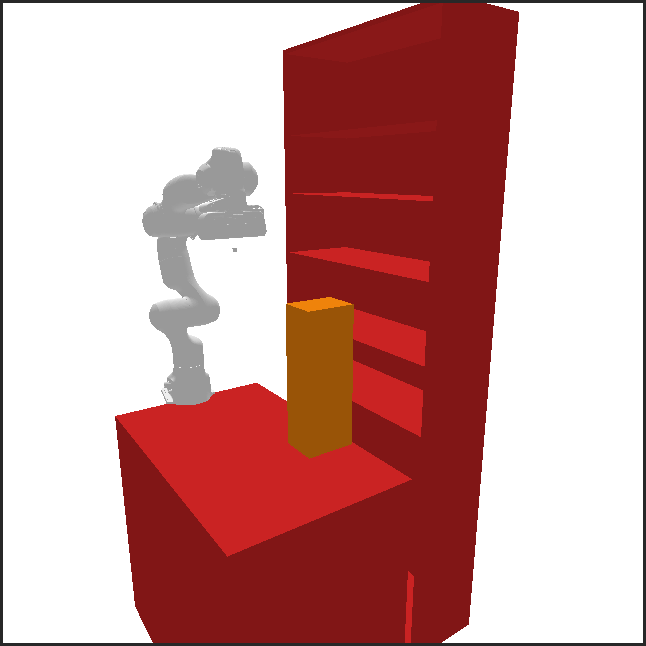}
    \caption{Tall table pillar}
    \label{fig:shelf-tall-table-pillar}
  \end{subfigure}
  \hfill
  \begin{subfigure}{0.19\textwidth}
    \centering
    \includegraphics[width=\linewidth]{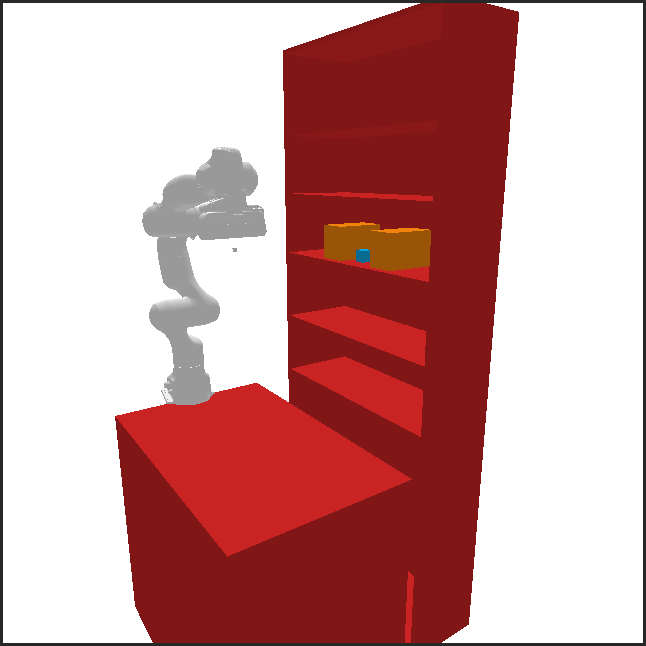}
    \caption{Two flankers (top)}
    \label{fig:shelf-two-flankers-top}
  \end{subfigure}
  \hfill
  \begin{subfigure}{0.19\textwidth}
    \centering
    \includegraphics[width=\linewidth]{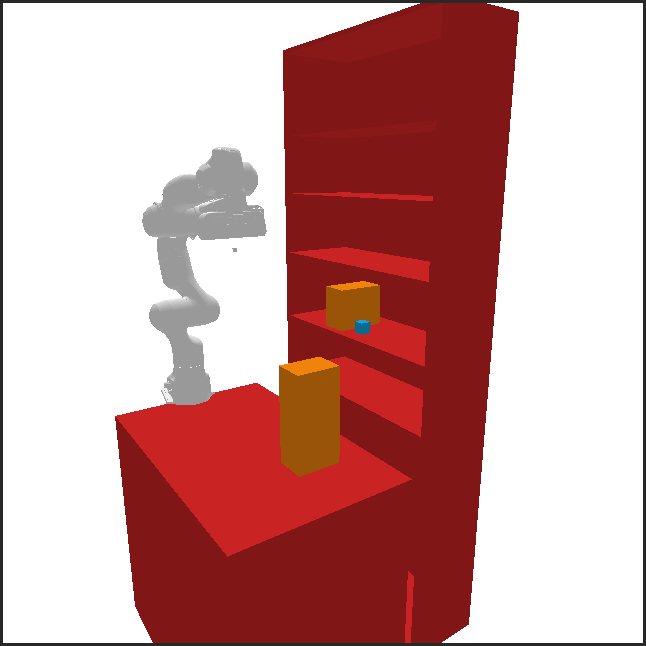}
    \caption{Mid-rung neighbor}
    \label{fig:shelf-mid-rung-neighbor-E}
  \end{subfigure}
  \hfill
  \begin{subfigure}{0.19\textwidth}
    \centering
    \includegraphics[width=\linewidth]{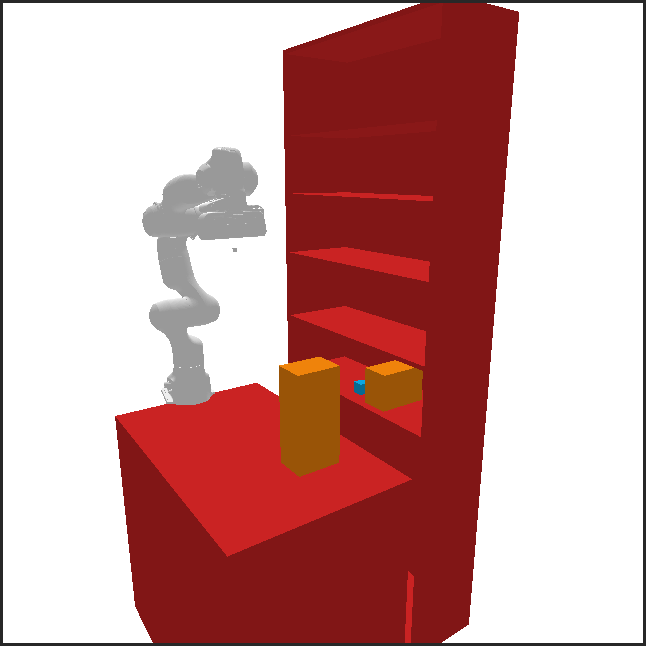}
    \caption{Low-rung neighbor}
    \label{fig:shelf-low-rung-neighbor-E}
  \end{subfigure}
  \caption{The five shelf scenarios (blue: cube to grasp, orange: obstacles).}
  \label{fig:shelf-scenarios}
\end{figure*}

\begin{table}[H]
\centering\small
\begin{tabular}{lcccccc}
\toprule
Method & Clutter & Pillar & Flankers & Mid-rung & Low-rung & Mean \\
\midrule
Unconstrained       & $3.3$  & $3.3$  & $3.3$  & $6.7$  & $0.0$  & $3.3$ \\
Gradient guidance   & $16.7$ & $70.0$ & $16.7$ & $33.3$ & $23.3$ & $32.0$ \\
Rejection           & $20.0$ & $83.3$ & $20.0$ & $33.3$ & $16.7$ & $34.7$ \\
Projection          & $40.0$ & $63.3$ & $46.7$ & $66.7$ & $60.0$ & $55.3$ \\
SafeDiffuser        & $6.7$  & $33.3$ & $3.3$  & $23.3$ & $0.0$  & $13.3$ \\
SafeFlowMatcher     & $43.3$ & $46.7$ & $53.3$ & $56.7$ & $23.3$ & $44.7$ \\
\AMR{}              & $\mathbf{80.0}$ & $\mathbf{93.3}$ & $\mathbf{83.3}$ & $\mathbf{76.7}$ & $\mathbf{73.3}$ & $\mathbf{81.3}$ \\
\bottomrule
\end{tabular}
\caption{\textbf{Shelf safe success (\%) per scenario}, $30$ tasks each.
\AMR{}'s mean logL per scenario: $19.29$ / $17.56$ / $17.63$ / $17.08$ / $18.26$.
Trust Sampling ($53.3\%$ overall) is not broken down by scenario.}
\label{tab:shelf-per-scenario}
\end{table}

\section{Additional 2D \textsc{Avoidance} results}\label{app:extra-2d}

\begin{figure}[H]
    \centering
    \includegraphics[width=0.9\linewidth]{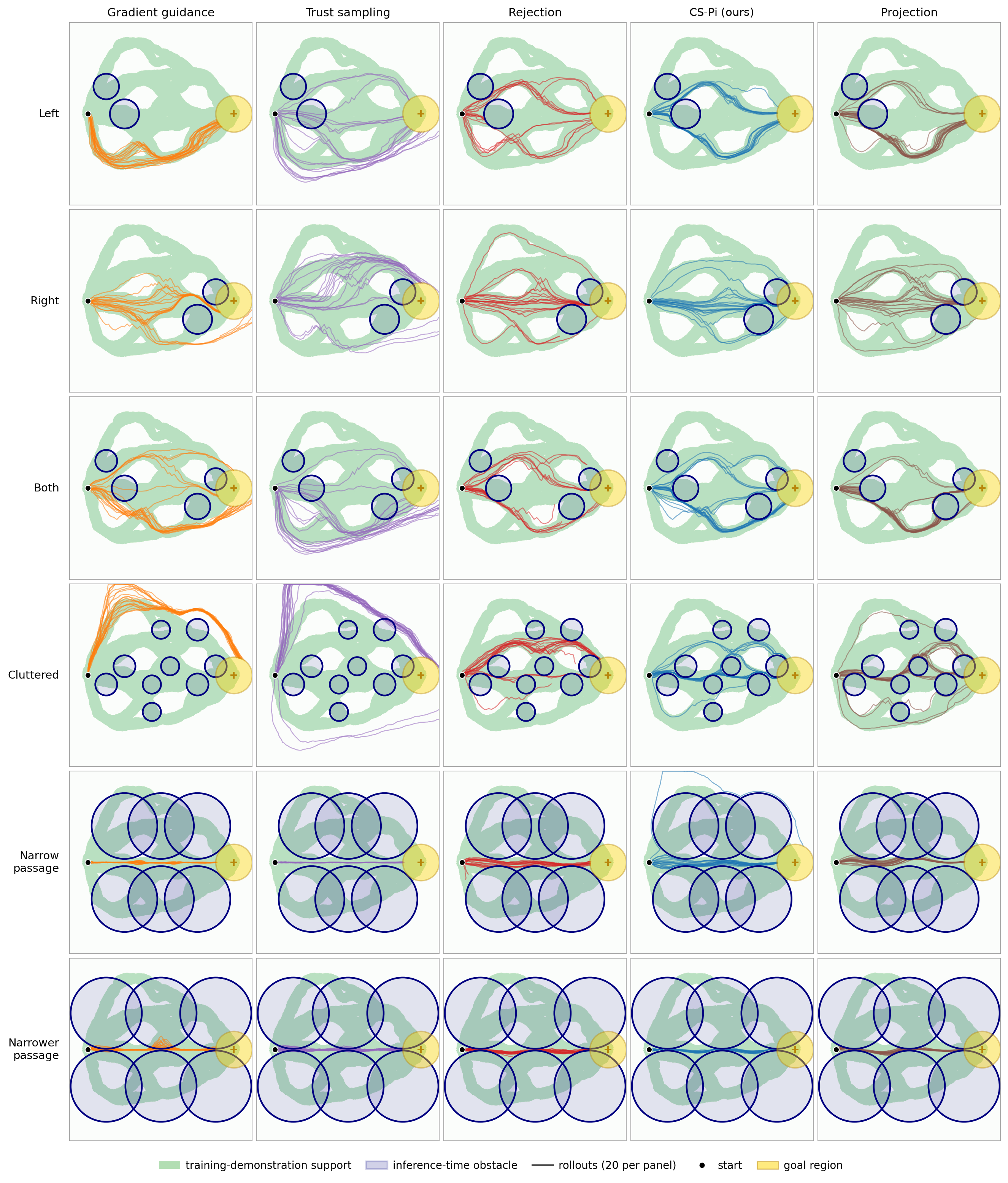}
    \caption{\textbf{2D \textsc{Avoidance} trajectory distributions, all six
    scenarios.} $20$ rollouts overlaid on the demonstration support (shaded);
    the obstacles are outlined.}
    \label{fig:2s_all}
\end{figure}

\begin{table}[H]
\centering\small
\begin{tabular}{lccccccc}
\toprule
Method & \texttt{left} & \texttt{right} & \texttt{both} & \texttt{clutt.} & \texttt{narrow} & \texttt{narrower} & Mean \\
\midrule
Unconstrained      & $20.0$ & $0.0$  & $0.0$  & $1.0$  & $10.0$ & $0.0$  & $5.2$ \\
Gradient guidance  & $99.5$ & $84.0$ & $15.0$ & $98.0$ & $100$  & $7.0$  & $67.2$ \\
Trust Sampling     & $9.0$  & $65.5$ & $1.0$  & $18.0$ & $99.0$ & $17.0$ & $34.9$ \\
Rejection          & $60.0$ & $95.0$ & $70.5$ & $71.5$ & $86.0$ & $93.5$ & $79.4$ \\
Projection         & $100$  & $100$  & $100$  & $100$  & $99.0$ & $100$  & $99.8$ \\
SafeDiffuser       & $100$  & $100$  & $100$  & $100$  & $100$  & $100$  & $100$ \\
SafeFlowMatcher    & $95.0$ & $77.5$ & $88.0$ & $10.5$ & $0.0$  & $0.0$  & $45.2$ \\
\bottomrule
\end{tabular}
\caption{\textbf{2D \textsc{Avoidance} safe success (\%) per scenario for the
baselines}, $200$ rollouts each (\AMR{}: $99.0\%$ overall, Table~\ref{tab:d3il-main}). Gradient guidance succeeds where one safe corridor dominates
(\texttt{left}, \texttt{cluttered}, \texttt{narrow}) and fails where the safe
set is a narrow passage the policy rarely uses (\texttt{both},
\texttt{narrower}).}
\label{tab:2d-per-scenario}
\end{table}

\begin{figure}[H]
    \centering
    \includegraphics[width=\linewidth]{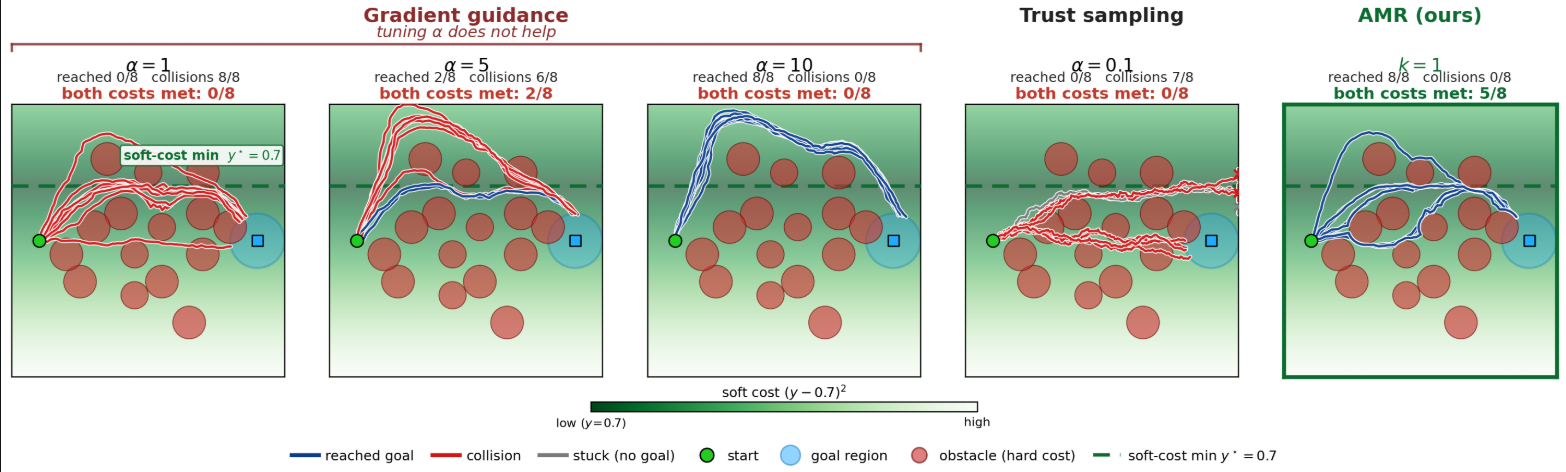}
    \caption{\textbf{Constraint and preference on 2D \textsc{Avoidance}
    (\texttt{cluttered}).} No setting of gradient guidance ($\alpha$) or Trust
    Sampling ($w$) satisfies both the obstacle constraint and the
    $(y-0.7)^2$ preference: low settings collide, high settings stay far from
    $y_{\mathrm{ref}}{=}0.7$. \AMR{} meets
    both.}
    \label{fig:softhard-2d}
\end{figure}

\paragraph{Composing a hard constraint with a soft preference.}
On the \texttt{cluttered} scenario we add to the obstacle constraint the
preference $L(\mathbf a)=(y-y_{\mathrm{ref}})^2$ with $y_{\mathrm{ref}}{=}0.7$.
\AMR{} applies the proximal step for the preference and then the projection, so
the constraint takes precedence; gradient guidance and Trust Sampling descend a
single weighted sum of the collision cost and the preference. A rollout
\emph{meets both} if it is collision-free and its mean height error
$\overline{|y_t-0.7|}$ is below $0.05$. \AMR{} meets both on every rollout for
$k\in\{30,100,200\}$; no setting of gradient guidance or Trust Sampling meets both
(Table~\ref{tab:softhard}, Fig.~\ref{fig:softhard-2d}): low settings collide and
high settings stay far from $y_{\mathrm{ref}}$. Projection alone and rejection
cannot target a preference and are omitted.

\begin{table}[H]
\centering\small
\begin{tabular}{llcccc}
\toprule
Method & Setting & Both met (\%) & Collision (\%) & Mean $|y-0.7|$ & NFE \\
\midrule
Unconstrained     & --               & $0$   & $35.3$ & $0.195$ & $10$ \\
\midrule
Gradient guidance & $\alpha{=}0.1$   & $0$   & $31.3$ & $0.194$ & $10$ \\
                  & $\alpha{=}0.3$   & $0$   & $28.0$ & $0.193$ & $10$ \\
                  & $\alpha{=}1$     & $0$   & $19.3$ & $0.189$ & $10$ \\
                  & $\alpha{=}3$     & $0$   & $0$    & $0.178$ & $10$ \\
                  & $\alpha{=}10$    & $0$   & $0$    & $0.150$ & $10$ \\
\midrule
Trust Sampling    & $w{=}0.01$       & $0$   & $25.3$ & $0.190$ & $70$ \\
                  & $w{=}0.05$       & $0$   & $0$    & $0.166$ & $70$ \\
                  & $w{=}0.1$        & $0$   & $0$    & $0.135$ & $70$ \\
\midrule
\AMR{}            & $k{=}1$          & $0$   & $0$    & $0.188$ & $110$ \\
                  & $k{=}10$         & $0$   & $0$    & $0.099$ & $82$ \\
                  & $k{=}20$         & $0$   & $0$    & $0.065$ & $63$ \\
                  & $k{=}30$         & $\mathbf{100}$ & $0$ & $0.043$ & $45$ \\
                  & $k{=}100$        & $\mathbf{100}$ & $0$ & $0.010$ & $23$ \\
                  & $k{=}200$        & $\mathbf{100}$ & $0$ & $\mathbf{0.007}$ & $21$ \\
\bottomrule
\end{tabular}
\caption{\textbf{Constraint and preference on 2D \textsc{Avoidance}
(\texttt{cluttered})}, $150$ chunks per setting.}
\label{tab:softhard}
\end{table}

\section{Why gradient guidance fails}\label{app:guidance-fail}
Gradient guidance steers a flow-matching policy toward a cost $L$ defined on
the \emph{clean} action $a_1$, such as a collision cost or an orientation
error. During sampling the integrator only ever holds a partially denoised
state $a_t$ at time $t\in(0,1)$, and the cost of a noisy state carries no
information about the cost of the clean action it will become: $L(a_t)$ and
$L(a_1)$ are unrelated. Guidance therefore cannot evaluate the cost at $a_t$.
It must instead predict the clean action the current state will denoise into
and evaluate the cost there. This requirement, forced by the cost being defined
on $a_1$, is the root of the failure.

\paragraph{The guidance direction one would want.}
Guidance aims to sample from the tilted distribution
$p^\star(a)\propto\pth(a)\,\exp(-\beta L(a))$, which concentrates the policy
on low-cost actions with strength $\beta$. The exact correction this adds
to the velocity at $a_t$ is the score of the tilt,
$\nabla_{a_t}\log\E_{p(a_1\mid a_t)}[\exp(-\beta L(a_1))]$. To first order
in $\beta$ this reduces to the posterior-mean gradient
\begin{equation}
  g^\star(a_t)\;=\;\E_{p(a_1\mid a_t)}\!\big[\nabla L(a_1)\big],
  \label{eq:ideal-guidance}
\end{equation}
the cost gradient averaged over \emph{every} clean action the noisy state
could still become, weighted by the denoising posterior $p(a_1\mid a_t)$.
Equation~\eqref{eq:ideal-guidance} is the guidance one would want: it
accounts for the policy's remaining uncertainty about $a_1$.

\paragraph{The Jensen swap.}
The expectation in Eq.~\eqref{eq:ideal-guidance} is intractable. The posterior
$p(a_1\mid a_t)$ is defined only implicitly through the flow and has no
closed form. Every gradient-guidance method makes the same substitution: it
moves the gradient outside the expectation,
\begin{equation}
  \E_{p(a_1\mid a_t)}\!\big[\nabla L(a_1)\big]
  \;\approx\;
  \nabla L\!\big(\E_{p(a_1\mid a_t)}[a_1]\big)
  \;=\;\nabla L(\bar a(a_t)),
  \label{eq:jensen-swap}
\end{equation}
replacing the expected gradient by the gradient at the expected action, the
proxy $\bar a(a_t)=\E[a_1\mid a_t]$. This is exact only when $\nabla L$ is
affine over the support of the posterior, that is, when $L$ is quadratic there.
For a general cost the two sides differ by the Jensen gap; a second-order
expansion of $L$ about $\bar a$ gives
\begin{equation}
  \underbrace{\E[L(a_1)]-L(\bar a)}_{\text{Jensen gap}}
  \;=\;\tfrac12\,\big\langle \nabla^2 L(\bar a),\,\mathrm{Cov}(a_1\mid a_t)\big\rangle+\dots,
  \label{eq:jensen-gap}
\end{equation}
and differentiating leaves a drift error of the same character: it scales with
the curvature of the cost and with the spread of the denoising posterior. Both
factors are largest exactly where guidance must act. The covariance
$\mathrm{Cov}(a_1\mid a_t)$ is wide at high noise, before the policy has
committed to an action; and a hard constraint is the limit of a cost that is
$0$ on $\C$ and unbounded off it, whose curvature diverges at the boundary.

\paragraph{The proxy is the Tweedie mean.}
The proxy $\bar a(a_t)=\E[a_1\mid a_t]$ is the posterior mean of the clean
action (the Tweedie estimate); for a flow policy it is recovered by a single
integration step from $a_t$ to $t=1$. When the posterior is unimodal and
narrow, the proxy is accurate and the swap is benign. When the policy is still
uncertain between \emph{distinct} clean actions, the posterior is multimodal
and its mean lies \emph{between} the modes, at an action the policy assigns
almost no mass and that need not be valid. The gradient $\nabla L(\bar a)$ is
then computed at a point on no mode. No choice of $\beta$ or step size repairs
this, because the error is in where the gradient is evaluated, not in its
magnitude.

\begin{figure}[H]
\centering
\includegraphics[width=0.9\textwidth]{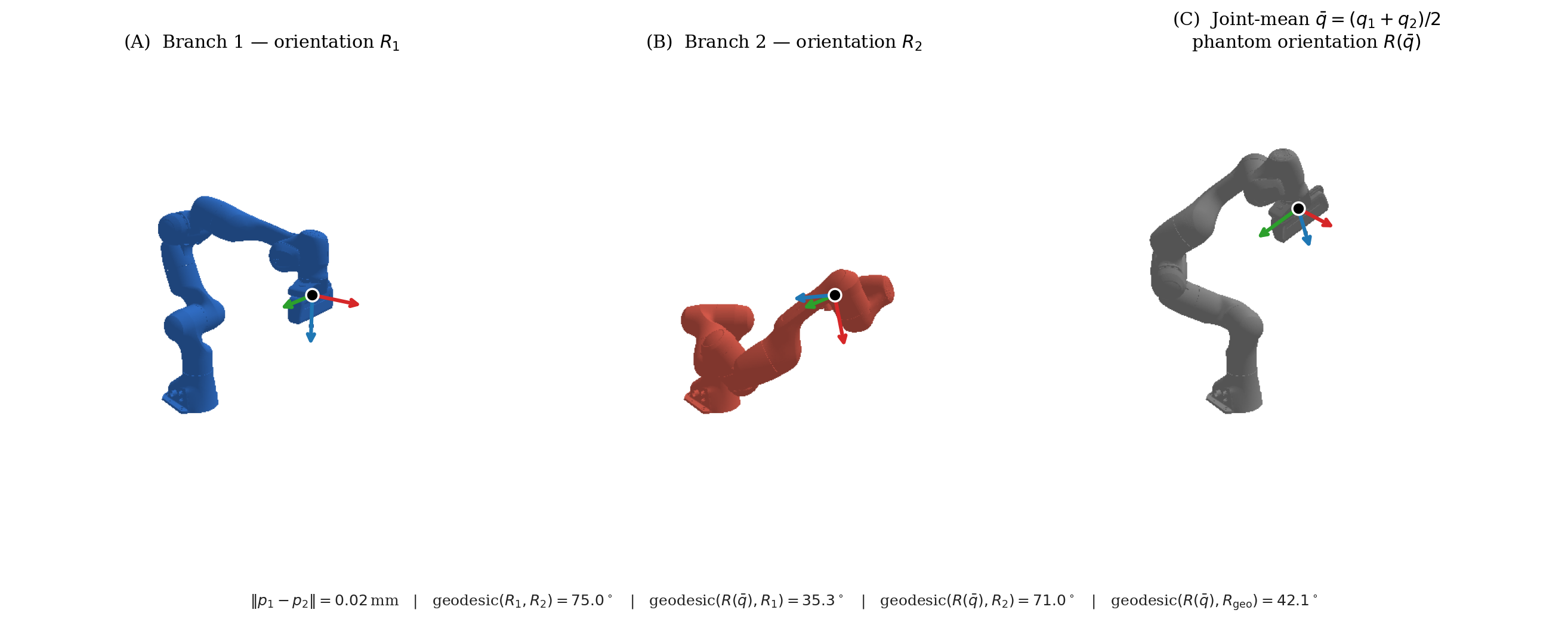}
\caption{\textbf{Saddle-ridge configuration of the orientation cost, rendered on
the Franka.} Inverse kinematics on a central target $p^*=(0.45,0.05,0.50)$.
(A), (B): two minima of $L$ at the same end-effector position ($0.02$\,mm
apart) with distinct joint configurations (blue: elbow up; red: elbow rotated
and wrist twisted). (C): the configuration $\bar q=(q_1+q_2)/2$ that gradient
guidance evaluates when its posterior places mass on both branches; the black
dot marks its end-effector position, $\mathbf{31}$\,cm from the target. The
average breaks both the position and the orientation target, although both
minima satisfy them.}
\label{fig:mpd9-saddle-ridge}
\end{figure}

\paragraph{\textsc{3D-Orient}: a worked failure.}
The orientation preference of Sec.~\ref{sec:exp-mpd9} is exactly this
multimodal case. The cost $L(q)=\lambda\lVert R(q)-R^\star\rVert_F^2$ on the
seven joint angles has several minima because of inverse-kinematics
redundancy: many distinct configurations reach the target orientation. When the
denoising posterior places mass on two such branches, the proxy is their
joint-space average, which lies between the basins and is not itself a valid
configuration; it breaks both the position and the orientation target.
Fig.~\ref{fig:mpd9-saddle-ridge} shows two minima $0.02$\,mm apart in
end-effector position whose joint-space mean lands $31$\,cm away. The gradient
there has no consistent relation to the gradient at either branch, so guidance
that descends it cannot honor the orientation without abandoning the reach
(Table~\ref{tab:mpd9}). \AMR{} evaluates the cost only on clean draws, each
inside a single basin.

\section{Exact per-iteration sampling law}\label{app:sampling-law}
This section makes precise the Gaussian-window interpretation of one \AMR{}
iteration. The model's conditional law admits the closed form below, whereas a
deterministic numerical flow defines a pushforward kernel whose density need
not equal that conditional law.

\paragraph{Setup.}
Let the flow-matching interpolant be
\begin{equation}
  \mathbf A_t=t\mathbf A_1+(1-t)\boldsymbol\varepsilon,\qquad
  \boldsymbol\varepsilon\sim\N(\mathbf{0},I),\quad
  \mathbf A_1\sim\pth ,
  \label{eq:fm-interpolant}
\end{equation}
so $p(\mathbf A_t\mid\mathbf A_1=\mathbf a)=\N(\mathbf A_t;t\mathbf a,(1-t)^2I)$.
At iteration $i$, define
\begin{equation}
  \mathbf a_c^{(i)}=\proj(\hat{\mathbf a}^{(i)}),\qquad
  \delta^{(i)}=\|\hat{\mathbf a}^{(i)}-\mathbf a_c^{(i)}\|,\qquad
  n^{(i)}=\frac{k}{k+\delta^{(i)}}.
  \label{eq:iter-indexed}
\end{equation}
For $\delta^{(i)}>0$, the rule for $n$ gives
\begin{equation}
  \frac{1-n^{(i)}}{n^{(i)}}=\frac{\delta^{(i)}}{k}=\sigma^{(i)}.
  \label{eq:window-width}
\end{equation}

\begin{proposition}[Conditional law of one iteration]
\label{prop:per-iteration-law}
Fix $\boldsymbol\mu^{(i)}\in\R^D$ and form
\begin{equation}
  \mathbf a_n^{(i)}=n^{(i)}\mathbf a_c^{(i)}+(1-n^{(i)})\boldsymbol\mu^{(i)}.
  \label{eq:conditional-anchor}
\end{equation}
If re-denoising draws from the model conditional
$p(\mathbf A_1\mid\mathbf A_{n^{(i)}}=\mathbf a_n^{(i)})$, then the next draw has
density
\begin{equation}
  q^{(i)}(\mathbf a)
  =\frac{1}{Z_i}\,\pth(\mathbf a)
   \exp\!\left[-\frac{\|\mathbf a-\tilde{\mathbf a}_c^{(i)}\|^2}{2(\sigma^{(i)})^2}\right],
  \qquad
  \tilde{\mathbf a}_c^{(i)}=\mathbf a_c^{(i)}+\sigma^{(i)}\boldsymbol\mu^{(i)},
  \label{eq:exact-window-law}
\end{equation}
where $Z_i$ is independent of $\mathbf a$: the policy seen through a Gaussian
window with standard deviation $\sigma^{(i)}=\delta^{(i)}/k$.
\end{proposition}

\begin{proof}
Bayes' rule and Eq.~\eqref{eq:fm-interpolant} give
\begin{align}
q^{(i)}(\mathbf a)
&\propto \pth(\mathbf a)\,p(\mathbf a_n^{(i)}\mid\mathbf A_1=\mathbf a)
 \propto \pth(\mathbf a)\,
  \exp\!\left(-\frac{\|\mathbf a_n^{(i)}-n^{(i)}\mathbf a\|^2}{2(1-n^{(i)})^2}\right).
\end{align}
Using Eq.~\eqref{eq:conditional-anchor},
$\mathbf a_n^{(i)}-n^{(i)}\mathbf a
=n^{(i)}\big(\mathbf a_c^{(i)}+\tfrac{1-n^{(i)}}{n^{(i)}}\boldsymbol\mu^{(i)}-\mathbf a\big)$.
Substituting Eq.~\eqref{eq:window-width} yields Eq.~\eqref{eq:exact-window-law}.
\end{proof}

\paragraph{Center and endpoint cases.}
For $\boldsymbol\mu^{(i)}=\mathbf{0}$ the window is centered on the projection
$\mathbf a_c^{(i)}$. Algorithm~\ref{alg:amr} uses
$\boldsymbol\mu^{(i)}=\boldsymbol\epsilon^{(i)}\sim\N(\mathbf{0},I)$; conditional
on that draw, Proposition~\ref{prop:per-iteration-law} remains exact with center
$\mathbf a_c^{(i)}+\sigma^{(i)}\boldsymbol\epsilon^{(i)}$. At $n=0$ the iteration
is an unconditional redraw, as in rejection sampling; at $n=1$ it returns the
anchor, as in projection.

\paragraph{Scope.}
Proposition~\ref{prop:per-iteration-law} is a per-iteration statement. It
assumes exact posterior re-denoising; it does not establish that deterministic
ODE integration samples this conditional, nor does it imply a stationary
distribution, a convergence rate, or a bound after several iterations.
Empirically, the projection residual decreases across iterations
(App.~\ref{app:nproj}).

\section{Hyperparameter sweeps and configuration selection}\label{app:hpsweep}
Every method in the main tables is reported at its best swept configuration.
\emph{logL} is the mean exact log-likelihood under $\pth$ (nats/dim);
\textsc{NFE} is the mean number of flow-model calls per rollout.

\paragraph{Swept hyperparameters.}
For \AMR{} we sweep $k$, the iteration cap $K_{\max}$ and the terminal projection
(on/off). For gradient guidance we sweep the step size $\alpha$. For Trust
Sampling we sweep the step size $w$, the maximum number of guidance steps per
denoising step $J$, the schedule (constant vs.\ linear), whether a projection is
applied after each step (``post-projection''), and, where a joint-limit penalty
applies, its weight.

\paragraph{3D \textsc{Sphere-Avoid}.}
Gradient guidance was swept over $\alpha\in[3\cdot10^{-4},0.3]$ (best
$\alpha{=}0.01$: $23.6\%$ / $4.7\%$) and Trust Sampling over $w\in[0.01,1.0]$
with constant and linear schedules (best $w{=}0.3$, $J{=}5$: $77.0\%$ /
$38.7\%$). Table~\ref{tab:hp-spheres} gives \AMR{} across $k$ ($K_{\max}{=}20$,
no early stopping). Success rises with $k$ and saturates on L1 by $k{=}20$.
Table~\ref{tab:spheres} uses $k{=}50$; $k{=}100$ is $4.3$ points higher on L2.

\begin{table}[H]
\centering\small
\begin{tabular}{lccccc}
\toprule
$k$ & $1$ & $10$ & $20$ & $50$ & $100$ \\
\midrule
Safe-SR L1 / L2 & $90.1$ / $59.8$ & $97.8$ / $77.7$ & $99.6$ / $88.3$ & $99.4$ / $90.2$ & $99.8$ / $94.5$ \\
NFE L1          & $72.0$ & $24.4$ & $20.9$ & $19.2$ & $18.7$ \\
\bottomrule
\end{tabular}
\caption{\textbf{\AMR{} on \textsc{Sphere-Avoid} across $k$} ($K_{\max}{=}20$, no
early stopping; the same evaluation as the no-stopping row of
Table~\ref{tab:es-spheres}). Table~\ref{tab:spheres} is a separate evaluation,
so its values at $k{=}50$ differ slightly.}
\label{tab:hp-spheres}
\end{table}

\paragraph{2D \textsc{Avoidance}.}
Gradient guidance uses the best $\alpha$ per scenario (macro $67.2\%$;
Table~\ref{tab:2d-per-scenario}); Trust Sampling reaches $34.9\%$.
\AMR{} reaches $99.2$--$99.3\%$ for $k\in\{5,10,20\}$
(Table~\ref{tab:schedule-ablation}).

\paragraph{Shelf.}
Table~\ref{tab:hp-shelf-trust} gives the full Trust Sampling sweep. Without
post-projection Trust Sampling peaks at $53.3\%$ ($w{=}0.1$, $J{=}5$), the reported
configuration. With
post-projection it peaks at $43.3\%$; post-projection is a terminal projection,
which no baseline uses in the main tables. \AMR{} across $k$ ($K_{\max}{=}20$, no
early stopping): $70.7$ / $78.7$ / $80.0$ / $76.7$ / $69.3\%$ at
$k{=}5$ / $10$ / $20$ / $50$ / $100$ (NFE $398$ / $376$ / $353$ / $306$ / $290$);
Table~\ref{tab:shelf} uses $k{=}50$, $K_{\max}{=}5$ ($81.3\%$).

\begin{table}[H]
\centering\small
\begin{tabular}{lccc|ccc}
\toprule
& \multicolumn{3}{c|}{no post-projection} & \multicolumn{3}{c}{post-projection} \\
$w$ & $J{=}2$ & $J{=}5$ & $J{=}10$ & $J{=}2$ & $J{=}5$ & $J{=}10$ \\
\midrule
$0.005$ & $2.7$  & $6.0$           & $4.0$                         & $29.3$ & $30.7$ & $32.7$ \\
$0.01$  & $3.3$  & $4.0$           & $11.3$                        & $30.0$ & $33.3$ & $34.0$ \\
$0.05$  & $9.3$  & $38.7$          & --                            & $31.3$ & $36.0$ & $35.3$ \\
$0.1$   & $28.0$ & $\mathbf{53.3}$ & $51.3$                        & $33.3$ & $32.0$ & $43.3$ \\
\bottomrule
\end{tabular}
\caption{\textbf{Trust Sampling on Shelf: safe success (\%) over $w\times J$},
$150$ tasks per cell (--: not run). Bold: the reported configuration (logL $12.23$, NFE
$1104$).}
\label{tab:hp-shelf-trust}
\end{table}

\paragraph{3D-Orient.}
We report gradient guidance at two step sizes: $\alpha{=}0.003$ keeps most of
the reach ($70.0\%$) and $\alpha{=}0.03$ gives its best success ($6.7\%$); larger
and smaller values are worse. Trust Sampling uses $w{=}0.001$ with the
joint-limit penalty.
\AMR{} uses $k{=}10$, $K_{\max}{=}3$, $\lambda_{ee}{=}3000$, $\beta{=}10^6$.

\begin{figure}[tbp]
  \centering
  \includegraphics[width=\linewidth]{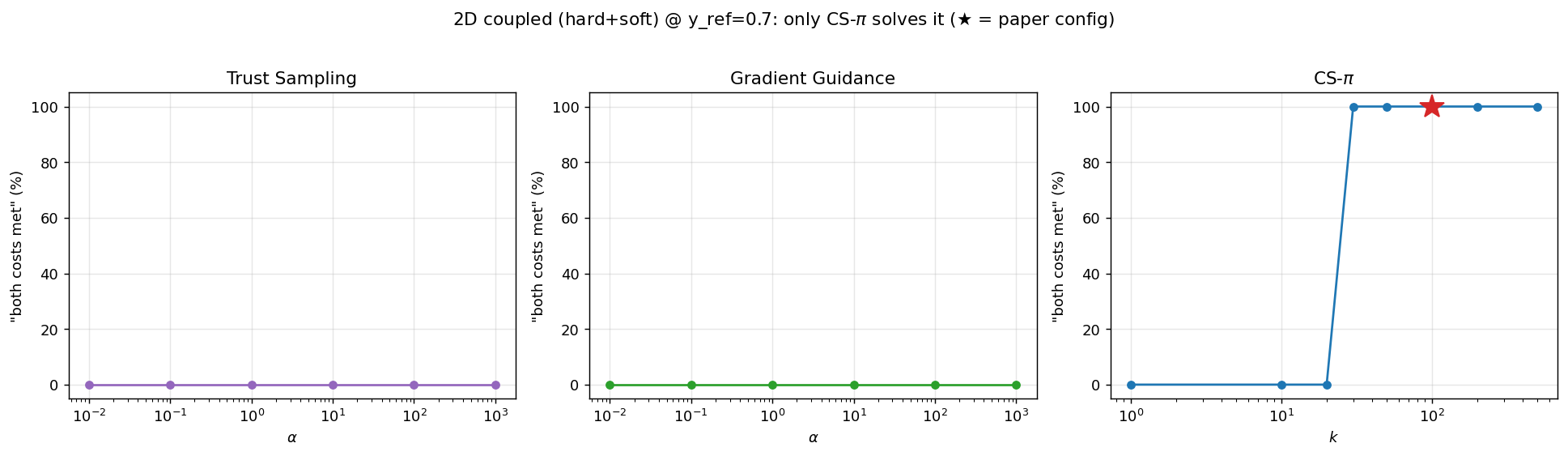}
  \caption{\textbf{Constraint and preference: sweep.} Fraction of rollouts
  meeting both for \AMR{} ($k$), gradient guidance ($\alpha$) and Trust Sampling
  ($w$); only \AMR{} meets both, for $k\in\{30,100,200\}$.}
  \label{fig:hp-coupled}
\end{figure}

\begin{figure}[tbp]
  \centering
  \begin{subfigure}{\linewidth}
    \centering
    \includegraphics[width=\linewidth]{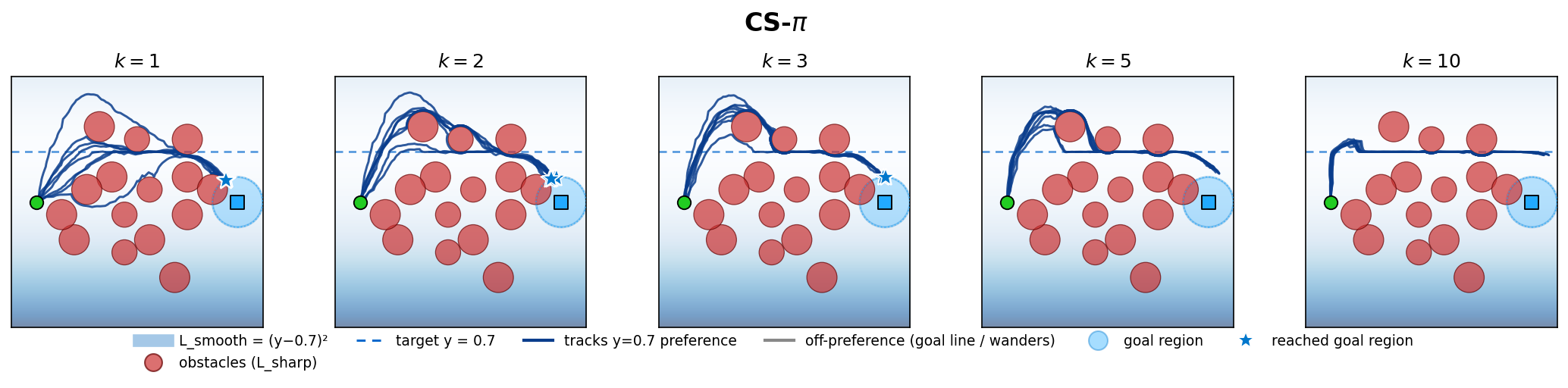}
    \caption{\AMR{} (proximal step, then projection)}
    \label{fig:coupled-traj-amr}
  \end{subfigure}\\[4pt]
  \begin{subfigure}{\linewidth}
    \centering
    \includegraphics[width=\linewidth]{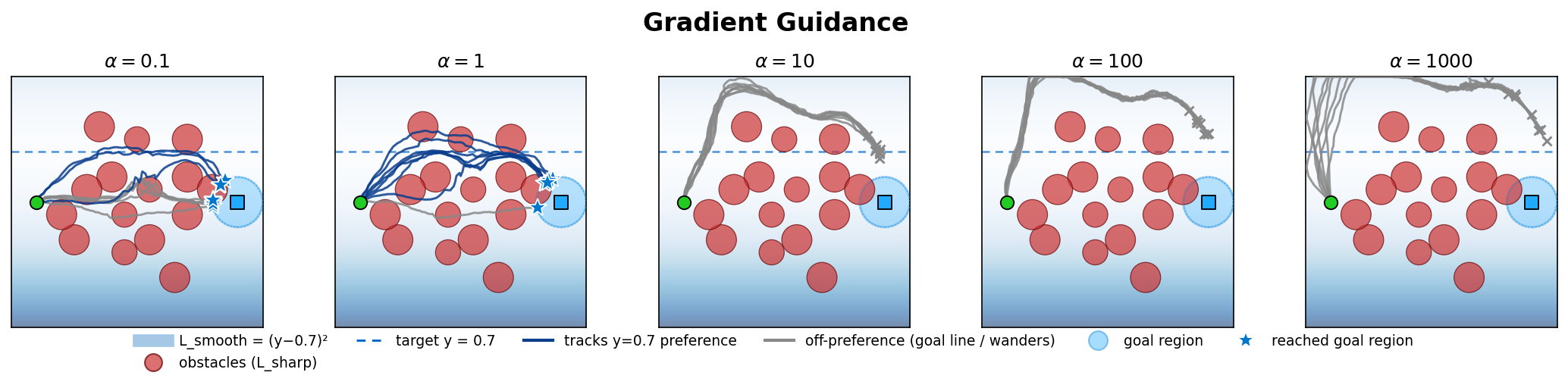}
    \caption{Gradient guidance}
    \label{fig:coupled-traj-tweedie}
  \end{subfigure}\\[4pt]
  \begin{subfigure}{\linewidth}
    \centering
    \includegraphics[width=\linewidth]{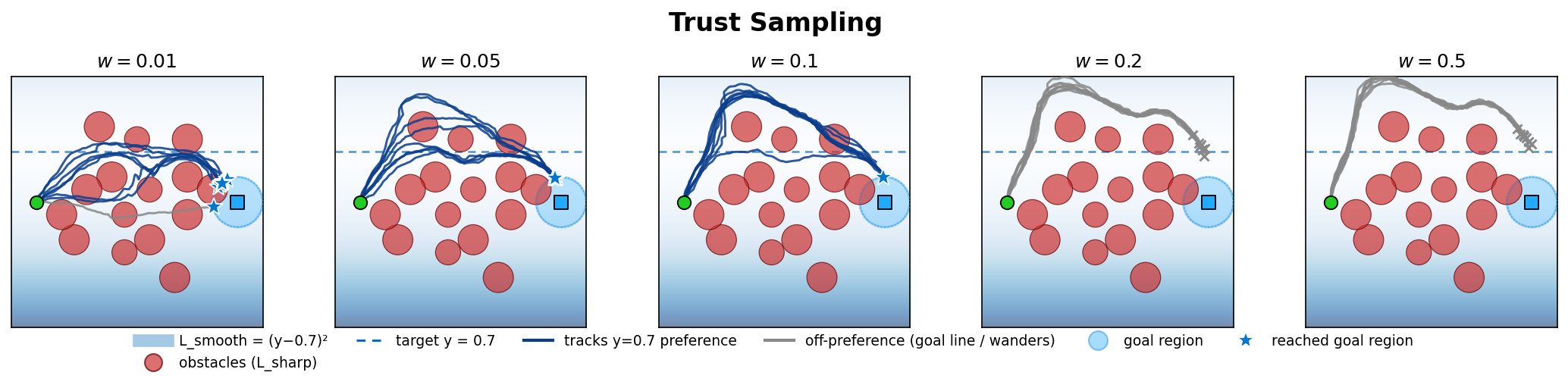}
    \caption{Trust Sampling}
    \label{fig:coupled-traj-trust}
  \end{subfigure}
  \caption{\textbf{Trajectories under the constraint and the $(y-0.7)^2$
  preference}, five settings per method. Only \AMR{}'s sweep contains settings
  that satisfy both; guidance and Trust Sampling trade one for the other at
  every setting.}
  \label{fig:coupled-traj}
\end{figure}

\section{Empirical convergence: time $n$ and projection residual}\label{app:nproj}
A successful \AMR{} loop drives the projection residual $\delta$ to zero, and
therefore $n=k/(k+\delta)$ to $1$. We log both per iteration
(Table~\ref{tab:nproj}; for \textsc{3D-Orient} and the preference, $\delta$ is
the proximal-step residual). Means are over loops still running at that
iteration; for 2D we give the first and last iteration per group of scenarios.

\begin{table}[H]
\centering\small
\begin{tabular}{llll}
\toprule
Benchmark ($k$) & loops & mean $n$ per iteration & residual first $\to$ last \\
\midrule
2D \texttt{left}/\texttt{right}/\texttt{narrow} ($10$) & -- & $0.64$--$0.81\to0.99$+ & $2.7$--$6.3\to\approx0.1$ \\
2D \texttt{both}/\texttt{cluttered} ($10$)     & -- & $0.63$--$0.68\to0.98$    & $6$--$8\to0.5$--$0.7$ \\
2D \texttt{narrower} ($10$)                     & -- & $0.73\to0.80$ (stalls)   & $4.1\to2.9$ \\
2D constraint + preference ($100$)      & $150$  & $0.743, 0.881, 0.919, 0.936, 0.946, 0.953$ & $34.6\to3.42$ \\
\textsc{Sphere-Avoid} L1 ($50$)         & $120$  & $0.991, 0.997, 0.999, 0.999, 0.999, 0.999$ & $0.478\to0.0071$ \\
\textsc{Sphere-Avoid} L2 ($50$)         & $120$  & $0.985, 0.993, 0.995, 0.997, 0.997, 0.997$ & $0.767\to0.0497$ \\
\textsc{Sphere-Avoid} L1 ($10$)         & $119$  & $0.955, 0.977, 0.986, 0.991, 0.993, 0.994$ & $0.467\to0.0225$ \\
Shelf ($50$)                            & $1508$ & $0.994, 0.998, 0.999, 0.999, 0.999$        & $0.287\to0.0376$ \\
Shelf ($1$)                             & $1602$ & $0.808, 0.818, 0.817, 0.817, 0.817$        & $0.237\to0.224$ \\
\textsc{3D-Orient} ($10$)               & $1440$ & $0.940, 0.966, 0.978$                       & $0.633\to0.222$ \\
\textsc{3D-Orient} ($1$)                & $1440$ & $0.633, 0.667, 0.673$                       & $0.579\to0.487$ \\
\bottomrule
\end{tabular}
\caption{\textbf{Time $n$ and projection residual per iteration.}}
\label{tab:nproj}
\end{table}

\paragraph{Where $n$ approaches 1.} On \textsc{Sphere-Avoid} and Shelf at their
chosen $k$ the residual falls $7\times$ (Shelf) to $67\times$ (L1). Because the
residuals are small relative to $k$, $n$ is close to $1$ from the first
iteration: the loop re-runs the policy from a barely noised anchor, so each draw
stays close to the projection. Lowering $k$ to $10$ on \textsc{Sphere-Avoid}
lowers the first-iteration $n$ to $0.955$ and the loop still converges.

\paragraph{Where it does not.} With $k{=}1$ on Shelf and \textsc{3D-Orient} the
loop stalls ($n$ stays at $0.82$ and $0.67$; the residual barely falls). On
Shelf this costs safe success ($66.7$ vs.\ $81.3\%$); on \textsc{3D-Orient} it
costs NFE ($348$ vs.\ $210$) and log-likelihood ($9.53$ vs.\ $10.21$) but not
success ($64.4$ vs.\ $65.6\%$). Under the preference the
residual falls tenfold but never reaches zero: the proximal step always has a
lower cost to move toward, so the loop refines rather than terminates.
\textsc{3D-Orient} keeps a small residual ($0.22$) because the orientation target
is never met exactly. On the easy 2D scenarios $n$ rises above $0.99$ within about four iterations and
the residual falls about $100\times$; on \texttt{narrower}, the hardest scene, $n$
rises only from $0.73$ to $0.80$ and the residual stalls ($4.1\to2.9$).

\section{Early stopping}\label{app:earlystop}
When the residual stops improving, further iterations cost compute without
helping. We compare four stopping rules: none; stop when $n>0.8$; stop when
$n>0.9$; and \emph{patience}, which stops when the last five iterations close
less than $5\%$ of the remaining gap between $n$ and $1$
(Tables~\ref{tab:es-spheres}--\ref{tab:es-shelf}; $K_{\max}{=}20$).

The threshold rules are tuned to 2D and do not transfer. On
\textsc{Sphere-Avoid} and Shelf, $n$ is close to $1$ from the first iteration at
moderate and large $k$, so these rules stop almost immediately and \AMR{}
performs close to projection ($88.2$ / $62.1\%$ on \textsc{Sphere-Avoid},
$55$--$63\%$ on Shelf). The patience rule is a trade-off rather than a free
saving. Patience matches no stopping on Shelf ($80.7$ vs.\ $80.0\%$ at $k{=}20$)
at $13\%$ fewer NFE, and on \textsc{Sphere-Avoid} L1 ($99.5$ vs.\ $99.4\%$ at
$k{=}50$) at slightly lower NFE, but costs $6$ points on L2 ($84.0$ vs.\
$90.2\%$). On 2D (Table~\ref{tab:es-2d}; means over $11$ values of $k$ and the six
scenarios, $200$ rollouts each) the patience rule matches the best threshold rule
($0.868$ vs.\ $0.864$ success) at $36\%$ fewer NFE than it and $54\%$ fewer than no
stopping. It also removes an over-refinement failure: at high $k$ on
\texttt{narrower}, running without stopping collapses to $\sim0\%$, while the rule
holds $100\%$.

\begin{table}[H]
\centering\small
\begin{tabular}{lccc}
\toprule
Rule & Success & NFE & Iterations \\
\midrule
none       & $0.825$          & $447$          & $3.94$ \\
$n>0.8$    & $0.864$          & $322$          & $2.62$ \\
$n>0.9$    & $0.863$          & $336$          & $2.85$ \\
patience   & $\mathbf{0.868}$ & $\mathbf{206}$ & $\mathbf{1.59}$ \\
\bottomrule
\end{tabular}
\caption{\textbf{Early stopping on 2D \textsc{Avoidance}}, means over $11$ values
of $k$ and six scenarios ($K_{\max}{=}20$).}
\label{tab:es-2d}
\end{table}

\begin{table}[H]
\centering\small
\begin{tabular}{lccccc}
\toprule
Rule & $k{=}1$ & $k{=}10$ & $k{=}20$ & $k{=}50$ & $k{=}100$ \\
\midrule
none      & $90.1$/$59.8$ ($72$) & $97.8$/$77.7$ ($24$) & $99.6$/$88.3$ ($21$) & $99.4$/$90.2$ ($19$) & $99.8$/$94.5$ ($19$) \\
$n>0.8$   & $90.4$/$62.9$ ($54$) & $88.2$/$62.1$ ($10$) & $88.2$/$62.1$ ($10$) & $88.2$/$62.1$ ($10$) & $88.2$/$62.1$ ($10$) \\
$n>0.9$   & $91.2$/$64.5$ ($68$) & $88.2$/$60.2$ ($10$) & $88.2$/$62.1$ ($10$) & $88.2$/$62.1$ ($10$) & $88.2$/$62.1$ ($10$) \\
patience  & $89.5$/$64.5$ ($31$) & $96.6$/$75.8$ ($22$) & $98.3$/$81.6$ ($20$) & $99.5$/$84.0$ ($18$) & $99.2$/$88.3$ ($17$) \\
\bottomrule
\end{tabular}
\caption{\textbf{Early stopping on \textsc{Sphere-Avoid}}: safe success L1 / L2
(\%) and, in parentheses, L1 NFE.}
\label{tab:es-spheres}
\end{table}

\begin{table}[H]
\centering\small
\begin{tabular}{lccccc}
\toprule
Rule & $k{=}5$ & $k{=}10$ & $k{=}20$ & $k{=}50$ & $k{=}100$ \\
\midrule
none      & $70.7$ ($398$) & $78.7$ ($376$) & $80.0$ ($353$) & $76.7$ ($306$) & $69.3$ ($290$) \\
$n>0.8$   & $56.0$ ($255$) & $56.0$ ($240$) & $55.3$ ($240$) & $56.7$ ($240$) & $55.3$ ($240$) \\
$n>0.9$   & $62.7$ ($272$) & $62.0$ ($253$) & $58.0$ ($240$) & $57.3$ ($240$) & $57.3$ ($240$) \\
patience  & $72.7$ ($312$) & $76.0$ ($307$) & $80.7$ ($307$) & $79.3$ ($291$) & $70.7$ ($280$) \\
\bottomrule
\end{tabular}
\caption{\textbf{Early stopping on Shelf}: safe success (\%) and, in
parentheses, NFE.}
\label{tab:es-shelf}
\end{table}

\section{Terminal projection: guarantee, how often it fires, and its cost}
\label{app:executed-guarantee}
\paragraph{Guarantee and per-benchmark use.}
With the terminal projection in Algorithm~\ref{alg:amr}, every executed action
lies in $\C$, provided $\C$ is nonempty and $\proj$ is exact. This is the
projection's guarantee, not one supplied by the loop; the loop's role is to
reduce the residual before that step. We use the terminal projection on 2D
\textsc{Avoidance}, \textsc{Sphere-Avoid}, Shelf and hardware, and not on
\textsc{3D-Orient}, where the target is a preference. No baseline uses one.

\paragraph{On versus off.} Table~\ref{tab:tp-onoff} runs \AMR{} with and without
the terminal projection. Without it, \AMR{} still has the highest safe success
on Shelf ($74.0\%$, against $55.3\%$ for projection and $53.3\%$ for Trust
Sampling) and on \textsc{Sphere-Avoid} L2 ($69.9\%$, against $62.1\%$ and
$38.7\%$); log-likelihood barely changes on these benchmarks.

\begin{table}[H]
\centering\small
\begin{tabular}{llccccc}
\toprule
Benchmark & Terminal proj. & Safe-SR & logL mean & logL median & Fire rate & NFE \\
\midrule
\textsc{Sphere-Avoid} L1 & on  & $99.9$ & $2.87$ & $2.81$ & $6.7\%$  & $20$ \\
                  & off & $98.1$ & $2.87$ & $2.81$ & --       & $20$ \\
\textsc{Sphere-Avoid} L2 & on  & $91.8$ & $2.54$ & $2.27$ & $46.9\%$ & $26$ \\
                  & off & $69.9$ & $2.55$ & $2.27$ & --       & $26$ \\
Shelf             & on  & $81.3$ & $17.96$ & $9.68$ & $14.7\%$ & $271$ \\
                  & off & $74.0$ & $17.59$ & $9.94$ & --       & $270$ \\
\bottomrule
\end{tabular}
\caption{\textbf{\AMR{} with and without the terminal projection.} Fire rate:
fraction of sampler calls still outside the tolerance after $K_{\max}$
iterations (Shelf: per plan step; \textsc{Sphere-Avoid}: per task, over the tasks with a
safe goal configuration).}
\label{tab:tp-onoff}
\end{table}

\paragraph{Log-likelihood of the chunks where it fires.}
Table~\ref{tab:fb} splits \AMR{}'s chunks by whether the terminal projection
fired. On \textsc{Sphere-Avoid} L1 and Shelf, the chunks where it fired keep a
median log-likelihood at or above one-shot projection's
($2.90$ vs.\ $2.41$; $14.70$ vs.\ $8.45$); on L2 they match it ($2.11$ vs.\
$2.13$). These chunks are the hardest subset by construction, whereas the
projection column is the median over all of projection's chunks.

On hardware the terminal projection fires on $91$ of $1879$ chunks ($4.8\%$;
$2.3\%$ in layout 1, $6.2\%$ in layout 2; in $15$ of $20$ trials), and those
chunks score lower than projection ($-518.0$ vs.\ $-473.2$). The raw policy
draws in these chunks are typical (median $-463.1$, against $-464.2$ for chunks
that converged), so the drop comes from the terminal projection rather than from atypical draws. Before
it fires, the loop has removed about $41\%$ of the initial violation; the
projection then moves the chunk by a median of $13.7$\,mm of end-effector motion
($0.56\times$ the chunk's own motion, $1.6\%$ of the workspace; p95 $170$\,mm),
and leaves a median residual penetration of $9.7$\,mm (p95 $40.3$\,mm), so on
hardware the terminal projection at the iteration cap does not certify
feasibility. Chunks that converged execute an unmodified policy sample with
median residual penetration $0.0$\,mm (p95 $0.76$\,mm).

\begin{table}[H]
\centering\small
\begin{tabular}{lccccc}
\toprule
Benchmark & Fires & Converged & Fired & Projection & Unconstrained \\
\midrule
\textsc{Sphere-Avoid} L1 & $6.7\%$  & $2.81$   & $2.90$   & $2.41$   & $2.74$ \\
\textsc{Sphere-Avoid} L2 & $46.9\%$ & $2.43$   & $2.11$   & $2.13$   & $2.90$ \\
Shelf                    & $14.7\%$ & $9.36$   & $14.70$  & $8.45$   & $10.00$ \\
Hardware                 & $4.8\%$  & $-469.1$ & $-518.0$ & $-473.2$ & $-470.2$ \\
\bottomrule
\end{tabular}
\caption{\textbf{Chunks where the terminal projection fires.} Fire rate and
median logL of converged and fired chunks, with projection and the
unconstrained policy as references. Medians, because a few fired chunks
dominate the mean.}
\label{tab:fb}
\end{table}

\paragraph{When $\C$ misses the policy's support.}
If $\C$ is nonempty but the policy assigns it negligible mass, no sampler can
return an action that is both safe and plausible. \AMR{} then degrades toward
its terminal projection. A residual that stops shrinking, equivalently an $n$
that stops rising, is a runtime signal of this case; App.~\ref{app:earlystop}
turns it into a stopping rule. On 2D \texttt{narrower} the residual falls from $4.1$ to $2.9$ and then stalls,
with $n$ stuck near $0.80$ (App.~\ref{app:nproj}).

\section{Compute}\label{app:walltime}
\textsc{NFE} counts flow-network calls but omits projection and proximal solves
and the backward passes of gradient guidance and Trust Sampling.
Table~\ref{tab:walltime} reports wall-clock time per rollout, measured on a
single machine. \AMR{} costs $2.1\times$ projection on \textsc{Sphere-Avoid}
and $1.6\times$ on Shelf, because of its projection and proximal solves. It is
$67\times$ faster than Trust Sampling on 2D, $1.2\times$ faster on Shelf, equal on
\textsc{3D-Orient}, and $2.1\times$ slower on \textsc{Sphere-Avoid}. The CBF
samplers are the most expensive on \textsc{Sphere-Avoid} and Shelf.
Hardware compute is in App.~\ref{app:hw-analysis}.

\begin{table}[H]
\centering\small
\begin{tabular}{lrrrr}
\toprule
Method & 2D & \textsc{Sphere-Avoid} & Shelf & \textsc{3D-Orient} \\
\midrule
Unconstrained      & $0.19$ & $0.19$  & $4.59$  & $0.47$ \\
Gradient guidance  & $0.52$ & $0.58$  & $22.1$  & $1.56$ \\
Trust Sampling     & $8.70$ & $2.02$  & $89.7$  & $12.48$ \\
Projection         & $0.08$ & $1.99$  & $49.0$  & -- \\
SafeDiffuser       & $0.13$ & $7.14$  & $164$   & -- \\
SafeFlowMatcher    & $0.16$ & $13.99$ & $167$   & -- \\
\AMR{}             & $0.13$ & $4.24$ & $76.0$ & $12.70$ \\
\bottomrule
\end{tabular}
\caption{\textbf{Wall-clock time per rollout (s)}, including flow calls,
backward passes, projection and proximal solves.}
\label{tab:walltime}
\end{table}

\section{Mechanism, sensitivity, and additional analyses}
\label{app:additional-analysis}

\paragraph{Why support membership can exceed the unconstrained policy's.}
Renoising and re-denoising pulls a sample toward high-density regions, as in
\textsc{SDEdit}, and feasibility and support membership are positively
correlated in these tasks, so stopping on feasibility further favors on-support
modes. Applying one renoise and re-denoise to an unmodified draw, with no constraint or
projection, raises the fully-on-support rate by up to $7.5$ points and the mean
fraction of on-support waypoints from $96.9\%$ to $98.6\%$.

\paragraph{Sensitivity to $k$.}\label{app:ktune}
Table~\ref{tab:fixedk} compares the tuned $k$ with a single fixed value for
every benchmark. A single $k{=}50$ is within $1.2$ points of the best $k$ on 2D,
\textsc{Sphere-Avoid} L1, Shelf and \textsc{3D-Orient}, and within $4.3$ points on
\textsc{Sphere-Avoid} L2. A single $k{=}20$ loses up to $7.4$ points
(\textsc{Sphere-Avoid} L2).
Within a benchmark, safe success varies by a few points over a broad range of
$k$ on \textsc{Sphere-Avoid} L1, Shelf and \textsc{3D-Orient}; \textsc{Sphere-Avoid}
L2 is more sensitive (Tables~\ref{tab:hp-spheres}, \ref{tab:es-shelf}).

\begin{table}[H]
\centering\small
\begin{tabular}{lcccc}
\toprule
Benchmark & Reported $k$ & Safe-SR (reported $k$) & Safe-SR ($k{=}20$) & Safe-SR ($k{=}50$) \\
\midrule
2D                       & $20$ & $99.0$ & $99.0$ & $99.1$ \\
\textsc{Sphere-Avoid} L1 & $50$ & $99.9$ & $99.2$ & $99.9$ \\
\textsc{Sphere-Avoid} L2 & $50$ & $91.8$ & $84.4$ & $91.8$ \\
Shelf                    & $50$ & $81.3$ & $74.7$ & $81.3$ \\
\textsc{3D-Orient}       & $10$ & $65.6$ & $64.4$ & $64.4$ \\
\bottomrule
\end{tabular}
\caption{\textbf{Reported versus fixed $k$.} Safe success (\%) at the $k$
reported in Tables~\ref{tab:d3il-main}, \ref{tab:shelf}, \ref{tab:mpd9} and
\ref{tab:spheres}, and at a single $k$ shared by every benchmark. Over all
sweeps, the best $k$ is within $1.2$ points of $k{=}50$ except on L2, where
$k{=}100$ is $4.3$ points higher (Table~\ref{tab:hp-spheres}). Shelf varies
most between evaluations: four further evaluations at $k{=}50$ give
$75.8\pm1.8$ (mean $\pm$ standard error; $78.7$ / $76.0$ / $78.0$ / $70.7$).
Each table is a separate evaluation, so values at the same $k$ can differ
slightly across tables.}
\label{tab:fixedk}
\end{table}

\paragraph{Schedule for $n$.}
Table~\ref{tab:schedule-ablation} compares the rule $n=k/(k+\delta)$ with a
constant $n$ and with renoising the projection with zero noise (adaptive $n$,
$\boldsymbol\epsilon=\mathbf 0$), on 2D \textsc{Avoidance} ($K_{\max}{=}5$,
$6\times200$ rollouts). The adaptive rule fires the terminal projection least
($8.7\%$ at $k{=}20$, against $15.6\%$ and $27.7\%$ for constant $n{=}0.8$ and
$0.5$). Zero-noise renoising matches the adaptive rule on fire rate and median
log-likelihood.

\begin{table}[H]
\centering\small
\begin{tabular}{lcccc}
\toprule
Schedule & Fire rate & Safe-SR & logL median & NFE \\
\midrule
adaptive, $k{=}5$ / $10$ / $20$      & $15.0$ / $9.5$ / $8.7\%$ & $99.2$ / $99.3$ / $99.3$ & $1.42$ / $0.38$ / $1.38$ & $180$ / $153$ / $145$ \\
constant $n{=}0.5$                   & $27.7\%$ & $99.1$ & $0.88$  & $181$ \\
constant $n{=}0.8$                   & $15.6\%$ & $99.6$ & $-2.23$ & $161$ \\
zero noise, $k{=}5$ / $10$ / $20$    & $15.0$ / $9.0$ / $8.6\%$ & $99.0$ / $99.6$ / $99.8$ & $1.16$ / $1.38$ / $1.42$ & $183$ / $150$ / $144$ \\
\bottomrule
\end{tabular}
\caption{\textbf{Schedule for $n$ on 2D \textsc{Avoidance}.} Fire rate is per
plan step. Medians, because a few chunks where the terminal projection fires
dominate the mean.}
\label{tab:schedule-ablation}
\end{table}

\paragraph{Smoothness.}
Table~\ref{tab:smooth} reports the per-rollout maximum acceleration and jerk
(second and third differences of the executed trajectory). On Shelf \AMR{} is
within $5\%$ of the unconstrained policy (acceleration $+4.5\%$, jerk $+4.8\%$),
and projection is the least smooth. On \textsc{Sphere-Avoid} \AMR{} is within
$1\%$ on L1 and $+14\%$ / $+11\%$ on L2, matching projection. On 2D the means
are similar across methods, but \AMR{} and projection have heavier
95th-percentile tails ($2.2$--$4.2\times$ the unconstrained policy). On
\textsc{3D-Orient} \AMR{}'s end-effector step matches the unconstrained policy
($0.07$\,m), but its joint-space acceleration is about $14\times$ and its jerk about
$11\times$ higher: the end-effector proximal step uses the arm's redundant joints. We
therefore claim end-effector, not joint-space, smoothness on \textsc{3D-Orient}.

\begin{table}[H]
\centering\small
\begin{tabular}{llcccc}
\toprule
& & \multicolumn{2}{c}{Acceleration} & \multicolumn{2}{c}{Jerk} \\
\cmidrule(lr){3-4}\cmidrule(lr){5-6}
Benchmark & Method & mean & p95 & mean & p95 \\
\midrule
Shelf & Unconstrained     & $0.518$ & $0.626$ & $0.841$ & $1.000$ \\
      & \AMR{}            & $0.541$ & $0.651$ & $0.882$ & $1.073$ \\
      & Projection        & $0.549$ & $0.666$ & $0.907$ & $1.097$ \\
      & Gradient guidance & $0.521$ & $0.619$ & $0.874$ & $1.013$ \\
      & Trust Sampling    & $0.524$ & $0.614$ & $0.878$ & $0.983$ \\
      & Rejection         & $0.521$ & $0.618$ & $0.845$ & $0.966$ \\
\midrule
\textsc{Sphere-Avoid} L1 & Unconstrained & $0.165$ & $0.243$ & $0.138$ & $0.221$ \\
      & \AMR{}            & $0.166$ & $0.241$ & $0.137$ & $0.215$ \\
      & Projection        & $0.171$ & $0.248$ & $0.142$ & $0.222$ \\
\midrule
\textsc{Sphere-Avoid} L2 & Unconstrained & $0.166$ & $0.216$ & $0.139$ & $0.196$ \\
      & \AMR{}            & $0.189$ & $0.247$ & $0.155$ & $0.217$ \\
      & Projection        & $0.192$ & $0.246$ & $0.155$ & $0.201$ \\
\midrule
2D    & Unconstrained     & $0.0058$ & $0.0088$ & $0.0061$ & $0.0092$ \\
      & \AMR{}            & $0.0068$ & $0.0196$ & $0.0090$ & $0.0306$ \\
      & Projection        & $0.0060$ & $0.0198$ & $0.0087$ & $0.0390$ \\
\midrule
\textsc{3D-Orient} & Unconstrained & $0.165$ & $0.204$ & $0.277$ & $0.347$ \\
      & \AMR{}            & $2.254$ & $4.574$ & $2.958$ & $8.812$ \\
      & Gradient guidance ($\alpha{=}0.03$) & $0.905$ & $1.499$ & $1.355$ & $2.667$ \\
\bottomrule
\end{tabular}
\caption{\textbf{Smoothness}: per-rollout maximum joint-space acceleration and
jerk, mean and 95th percentile over rollouts.}
\label{tab:smooth}
\end{table}

\paragraph{CBF certificates.}
A safety certificate alone does not make a sample useful: projection already
supplies one. Only SafeDiffuser matches \AMR{}, and only on 2D
(Table~\ref{tab:d3il-main}); both CBF samplers trail it on \textsc{Sphere-Avoid}
(Table~\ref{tab:spheres}) and Shelf (Table~\ref{tab:shelf}).
SafeFlowMatcher on Shelf uses $\rho{=}0.7$, $\epsilon{=}5$, $\alpha{=}0.07$,
$T_p{=}10$, $T_c{=}8$ (swept $\rho\in\{0.6,0.7,0.8\}$).

\section{Blocked-mode diagnostic}\label{app:blocked}
\begin{figure}[tbp]
\centering
\includegraphics[width=0.95\linewidth]{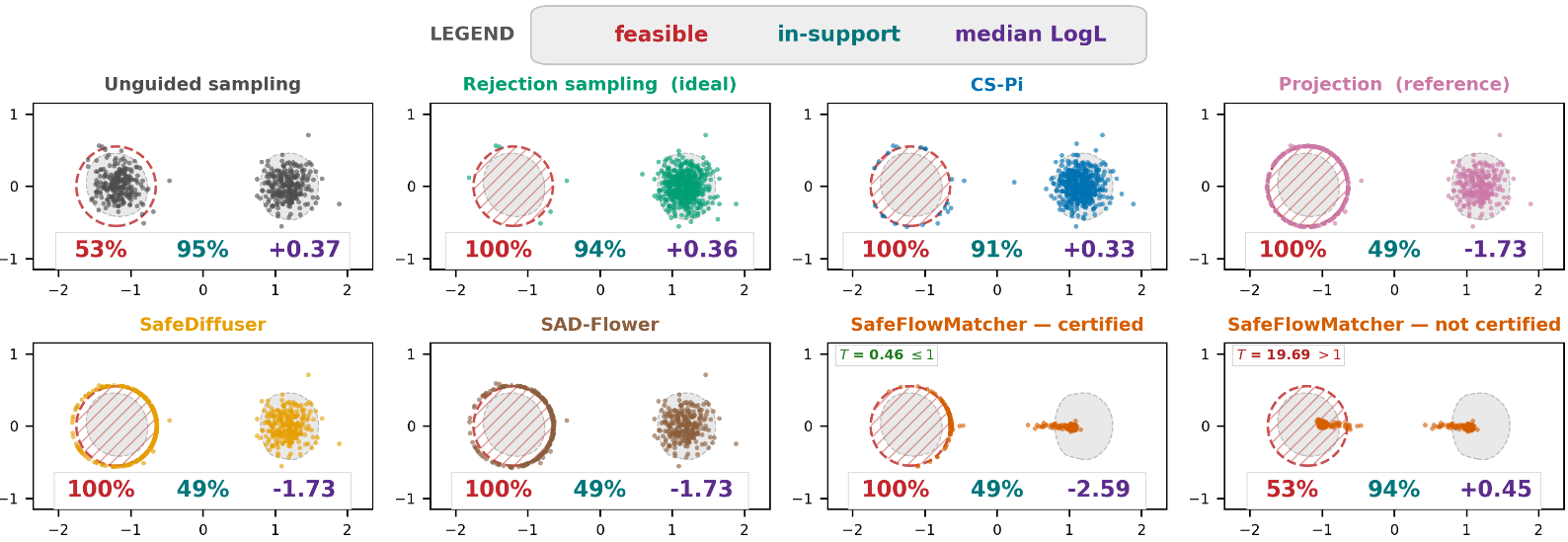}
\caption{\textbf{Blocked-mode diagnostic, left mode blocked.} Rejection and
\AMR{} move blocked draws to the surviving mode; the other methods stay safe but
land in low-likelihood regions of the policy.}
\label{fig:cbf-2d}
\end{figure}

This experiment isolates the claim in Sec.~\ref{sec:related} that a correction
applied during denoising can move a sample out of an unsafe region but cannot
move it to another mode of the policy. A flow-matching policy is trained on a
symmetric two-mode Gaussian mixture in 2D (modes at $(\pm1.2,0)$, standard
deviation $0.18$). A disk of radius $0.55$ centered on the left mode blocks it
entirely. Because the problem is 2D, exact log-likelihoods and exact rejection
sampling are tractable, so every method is compared with the correct answer:
the policy restricted to the safe set. We evaluate the \emph{blocked subset},
the $241$ of $500$ paired initial draws that land in the blocked mode
(Fig.~\ref{fig:cbf-2d}).

\begin{table}[H]
\centering\small
\begin{tabular}{lccc}
\toprule
Method & Moved to the surviving mode & median $\Delta\log p$ & NFE \\
\midrule
Rejection (exact reference)        & $92.5\%$ & $-0.03$ & $284$ \\
\AMR{}                             & $85.9\%$ & $-0.08$ & $282$ \\
Projection                         & $0.0\%$  & $-3.56$ & $100$ \\
SafeDiffuser ($\lambda{=}50$)      & $0.0\%$  & $-3.56$ & $100$ \\
SafeFlowMatcher (published defaults) & $0.0\%$ & $-6.42$ & $257$ \\
\bottomrule
\end{tabular}
\caption{\textbf{Blocked-mode diagnostic, blocked subset} ($241$ draws).
$\Delta\log p$: change in exact log-likelihood relative to the unconstrained
policy. Every method returns a safe action on all draws.}
\label{tab:blocked}
\end{table}

All methods are safe on every draw, so safety does not separate them
(Table~\ref{tab:blocked}). \AMR{} moves $85.9\%$ of blocked draws to the surviving
mode at a log-likelihood within $0.05$ nats of exact rejection and at the same
cost. Projection, SafeDiffuser and SafeFlowMatcher at its published defaults move
none: each applies a deterministic correction to the blocked sample, so its
output stays near the blocked mode. In a control where the constraint touches
neither mode, \AMR{} is statistically indistinguishable from the unconstrained
policy. SafeFlowMatcher was swept over $577$ configurations, SafeDiffuser over
$3$, and \AMR{} over $27$. One SafeFlowMatcher setting (a single prediction
step) does move most blocked draws, but it moves samples to one mode even when
the constraint touches neither mode, and returns unsafe actions when the policy
has a single mode, so the move is not driven by the constraint.

\section{Reliability of posterior-mean guidance and approximate projection}
\label{app:gradient-reliability}
The guidance proxy in Eq.~\eqref{eq:jensen-swap} can fail even for a unimodal
posterior. The relevant quantity is how quickly the cost curvature varies over
the posterior's spread.

\begin{proposition}[Posterior-mean gradient error]
\label{prop:gradient-reliability}
Let $p(a_1\mid a_t)$ have mean $\bar a$ and finite second moment. If $L$ has an
$L_H$-Lipschitz Hessian in operator norm on the relevant domain, then
\begin{equation}
\left\|\E_{p(a_1\mid a_t)}[\nabla L(a_1)]-\nabla L(\bar a)\right\|
\leq \frac{L_H}{2}\,\mathrm{tr}\,\mathrm{Cov}(a_1\mid a_t).
\label{eq:gradient-reliability-bound}
\end{equation}
\end{proposition}

\begin{proof}
Taylor expansion of $\nabla L$ about $\bar a$ gives
$\nabla L(a_1)=\nabla L(\bar a)+\nabla^2L(\bar a)(a_1-\bar a)+r(a_1)$ with
$\|r(a_1)\|\leq \frac{L_H}{2}\|a_1-\bar a\|^2$
\citep[Lemma~1.2.4]{nesterov2004introductory}. Taking expectations cancels the
linear term; then $\|\E[r]\|\le\E\|r\|\le (L_H/2)\E\|a_1-\bar a\|^2$ and
$\E\|a_1-\bar a\|^2=\mathrm{tr}\,\mathrm{Cov}(a_1\mid a_t)$.
\end{proof}

The proxy is therefore reliable when the curvature varies slowly at the scale of
the posterior's spread. Multimodality is not required for bias: a wide unimodal
posterior suffices. Multimodality creates the more severe case, in which the
posterior mean lies off-support between modes.

\paragraph{Approximate and non-convex projections.}
For non-convex $\C$ the projection can be set-valued, and iterative solvers may
return only a local or approximate solution. During \AMR{} iterations this point
is an anchor rather than the executed action, so renoising and re-denoising can
absorb anchor error and move to another mode. Projection has no such
correction. Only \AMR{}'s optional terminal step executes a projection; exact
safety of the executed action still requires that step to be exact.

\section{Benchmarks, metrics, and implementation details}\label{app:details}

\paragraph{Environments.}
\textbf{2D \textsc{Avoidance}} is a planar reaching task in $[0,1]^2$ with start
$(0.1,0.5)$ and goal $(0.9,0.5)$; the state is $(x,y,g_x,g_y)\in\R^4$ and the
action a velocity $(v_x,v_y)$, in chunks of $H{=}16$ steps with $4$ executed per
replan. Six training obstacles are present in every scenario; each test scenario
adds obstacles that close off some of the demonstrated paths (\texttt{left},
\texttt{right}, \texttt{both}, \texttt{cluttered}, \texttt{narrow},
\texttt{narrower}). The safe set is the collision-free set.
\textbf{3D \textsc{Sphere-Avoid}} (App.~\ref{app:sphere}) is a simulated 7-DoF
Franka Panda planning task in joint space, parameterized as a B-spline with
$28$ control points per joint; the safe set is the collision-free set of the
swept-sphere geometry plus joint limits. \textbf{Shelf}
(Sec.~\ref{sec:exp-shelf}) uses the same simulated arm for pick-and-place in five
scenarios (App.~\ref{app:shelf}); actions are chunks of $16$ absolute joint
configurations. \textbf{\textsc{3D-Orient}} is a 7-DoF reach to a goal anywhere
in the reachable workspace, with an orientation preference handled by a
proximal step on the end-effector orientation cost.

\paragraph{Metrics.}
\emph{Safe-SR} is the fraction of rollouts that complete the task and satisfy
the constraint (2D: reach the goal collision-free; \textsc{Sphere-Avoid}: reach
the goal with a collision-free trajectory; Shelf: pick and place the cube
collision-free). On \textsc{3D-Orient}, \emph{success} requires a step within
$10$\,cm of the target position and within $0.36$\,rad of the target
orientation. \emph{fully-in-supp.}\ is the fraction of rollouts whose every
waypoint lies inside a dilated convex hull of the demonstrations. \emph{logL} is
the exact log-likelihood under $\pth$ (nats/dim), averaged over chunks.
\textsc{NFE} is the mean number of flow-model calls per rollout.

\paragraph{Configuration.}
All simulated experiments use deterministic Euler integration.
Table~\ref{tab:impl} lists the per-benchmark settings.

\begin{table}[H]
\centering\small
\setlength{\tabcolsep}{4pt}
\begin{tabular}{llllll}
\toprule
Benchmark & Action & \AMR{} & Term.\ proj. & $\tau$ & Rollouts \\
\midrule
2D \textsc{Avoidance} & $(v_x,v_y)$, $H{=}16$    & $k{=}20$, $K_{\max}{=}5$  & on  & $0.01$  & $6\times200$ \\
\textsc{Sphere-Avoid} & B-spline, $28\times7$    & $k{=}50$, $K_{\max}{=}20$ & on  & $0.001$ & $120\times8$ \\
Shelf                 & joint chunk, $16\times7$ & $k{=}50$, $K_{\max}{=}5$  & on  & $0.001$ & $150$ ($30\times5$) \\
\textsc{3D-Orient}    & 7-DoF joints             & $k{=}10$, $K_{\max}{=}3$  & off & $0.001$ & $90$ ($30\times3$) \\
Real world            & joint velocities         & $k{=}5$, $K_{\max}{=}5$   & on  & $0.001$ & $20$ ($10\times2$) \\
\bottomrule
\end{tabular}
\caption{\textbf{Per-benchmark configuration.} $\tau$: stopping tolerance.
\textsc{Sphere-Avoid} rollouts: $120$ tasks $\times\,8$ samples.
\textsc{3D-Orient} also uses $\lambda_{ee}{=}3000$ and $\beta{=}10^6$.}
\label{tab:impl}
\end{table}

\section{Hardware deployment setup}\label{app:hardware}
We deploy \AMR{} zero-shot on a Franka FR3 with a Robotiq 2F-85 gripper,
controlled through FrankaPy at $15$\,Hz. The policy is $\pi_{0.5}$ finetuned on
DROID; it acts in joint-velocity space, so every Cartesian constraint is
enforced through forward kinematics and Jacobians on the integrated joint
trajectory $Q[t]=q_0+\Delta t\sum_{\le t}V$, with $\Delta t$ the control period.

\paragraph{Perception to constraints.}
The ZED 2 side camera (RGB and metric depth) drives the constraint pipeline;
the ZED mini wrist camera is used only by the policy. All geometry is expressed
in the robot base frame. For benchmarking we assume a static scene and run the
perception pipeline once per rollout: SAM2 segments the scene with no text
prompt, each mask is unprojected through depth into a point cloud, and a
vision-language model (Gemini) assigns only semantic roles (which segment to
grasp, which to place on, which to ignore). It never produces coordinates; all
positions and extents come from depth. Each obstacle is fit with a tight
oriented bounding box; the grasp and target objects reduce to centroids.
Because a single depth view sees only an object's front, we mirror the occluded
half across the camera-ray mid-plane before fitting, which enlarges the geometry
in the safe direction.

\paragraph{Constraints and execution.}
A task state machine (\textsc{reach}\,$\to$\,\textsc{grasp}\,$\to$\,%
\textsc{transport}\,$\to$\,\textsc{place}) rebuilds the active constraints every
chunk: a goal attractor at the grasp or target centroid, a collision constraint
between the robot and gripper sphere model and the obstacle boxes, a table
half-space, and, once grasped, a sphere for the held object. \AMR{}, gradient
guidance, Trust Sampling and projection receive the identical obstacle set,
$q_0$ and $\Delta t$, so the comparison differs only in the sampler. The
collision cost takes the form each method is designed for: a hinge penetration
penalty for \AMR{} and projection, and an OmniGuide-style log-barrier on
clearance for gradient guidance and Trust Sampling. The perceived scene is saved
to a scene file and replayed across methods.

\paragraph{Planned versus realized motion.}
Collisions are scored on the \emph{realized} joint configuration each tick with
zero safety margin. FrankaPy controller lag leaves the arm short of each
commanded waypoint ($\sim\!50$\,mm of end-effector travel), so a chunk with zero
planned constraint cost can still log a contact. This execution gap applies
equally to every method.

\subsection{Hardware execution analysis}\label{app:hw-analysis}
The following statistics are computed offline from the logged trials ($20$
\AMR{} trials, $1879$ chunks).

\paragraph{What executes.}
On $73.9\%$ of chunks the raw $\pi_{0.5}$ draw is already safe and executes
unchanged (bit-identical); on $21.2\%$ the loop converges and executes an
unmodified re-denoised policy sample; on $4.8\%$ the terminal projection fires
(App.~\ref{app:executed-guarantee}). Iterations per chunk: $1$ ($1389$ chunks),
$2$ ($198$), $3$ ($120$), $4$ ($56$), $5$ ($116$). The median ratio of
successive residuals is $0.20$, $0.33$, $0.61$ and $0.77$ over the first four
iterations ($96$, $91$, $81$ and $75\%$ of ratios below $1$): contraction is
strongest at the first iteration, and the violation that survives to the later
iterations is the one that ends in the terminal projection.

\paragraph{Smoothness.}
Table~\ref{tab:hw-jerk} reports the per-trial maximum end-effector jerk,
differentiated against measured timestamps. \AMR{}'s median is the highest
(the correction adds motion), but it has the lightest tail: the ratio of the
95th percentile to the median is $1.64$, against $1.80$ for the unconstrained
policy and $5.69$ for projection.

\begin{table}[H]
\centering\small
\begin{tabular}{lcccc}
\toprule
Method & Trials & Median & p95 & p95 / median \\
\midrule
Unconstrained      & $20$ & $116.9$ & $210.3$ & $1.80$ \\
Gradient guidance  & $20$ & $90.9$  & $322.7$ & $3.55$ \\
Trust Sampling     & $10$ & $178.8$ & $333.5$ & $1.87$ \\
Projection         & $17$ & $167.0$ & $949.7$ & $5.69$ \\
\AMR{}             & $20$ & $181.6$ & $298.7$ & $\mathbf{1.64}$ \\
\bottomrule
\end{tabular}
\caption{\textbf{Hardware smoothness}: per-trial maximum end-effector jerk
(m/s$^3$), median and 95th percentile across trials.}
\label{tab:hw-jerk}
\end{table}

\paragraph{Dynamic validity.}
The projection certifies joint-position limits, collision and the table, but not
velocity or acceleration limits. The realized joint velocity never exceeded the
Franka limits ($0$ of $130{,}361$ joint-ticks; worst $0.77\times$ the limit),
because the impedance controller tracks only part of each commanded step. The
commanded stream does exceed them occasionally: $0.08\%$ of joint-ticks over the
velocity limit within converged chunks, which execute unmodified policy samples,
and $2.0\%$ within chunks where the terminal projection fired. Adding velocity
and acceleration bounds to the projection would extend the certificate to them.

\paragraph{Compute.}
\AMR{} makes $1.62$ policy calls per chunk on average ($0.62$ re-denoises) and
$2.57$ projection solves; its mean NFE is $10.9$ against $10$ for a single draw,
because with $k{=}5$ the time $n$ is close to $1$ and a re-denoise needs about one
integration step. Per-step latency is reported in Table~\ref{tab:hw}.

\end{document}

%% file: intro_st.tex
\section{Introduction}
\label{sec:intro}
\vspace{-4pt}
Large-scale flow-matching and diffusion policies have become capable generalist priors for robot manipulation\citep{black2026pi0visionlanguageactionflowmodel,intelligence2025pi05visionlanguageactionmodelopenworld}: trained once on broad demonstrations, a single model can pick and place objects, open drawers, and fold clothes. Yet no training set anticipates every deployment, and such a policy will inevitably meet safety-critical scenarios its demonstrations never contained, for example: a wine glass left beside the drawer it must open, which the arm must not touch. This requirement is a \emph{constraint}, supplied only at deployment. It defines a \emph{safe set}: the set of actions that satisfy it; in our example, every action that does not touch the glass. Our goal is to sample actions that both come from the policy and lie in the safe set, without retraining the policy for each new constraint. The natural approach is to correct the policy's output: sample an action and project it onto the safe set, moving it to the nearest safe action, or steer the denoising trajectory using a function of the constraint. Such corrections address the second half of this goal, moving the action into the safe set. They can lose the first: the correction can move the action somewhere the policy would rarely sample. This matters because the policy is what captures the task. An action it considers unlikely may clear the wine glass yet no longer reach the drawer handle. Such an action also loses what the policy learned from the demonstrations, beyond the task itself: smooth, kinematically feasible motion. We therefore require the sampled action to be both \emph{safe}, satisfying the constraint, and \emph{plausible}, remaining likely under the policy.

\begin{figure}[t]
\centering
\includegraphics[width=\textwidth]{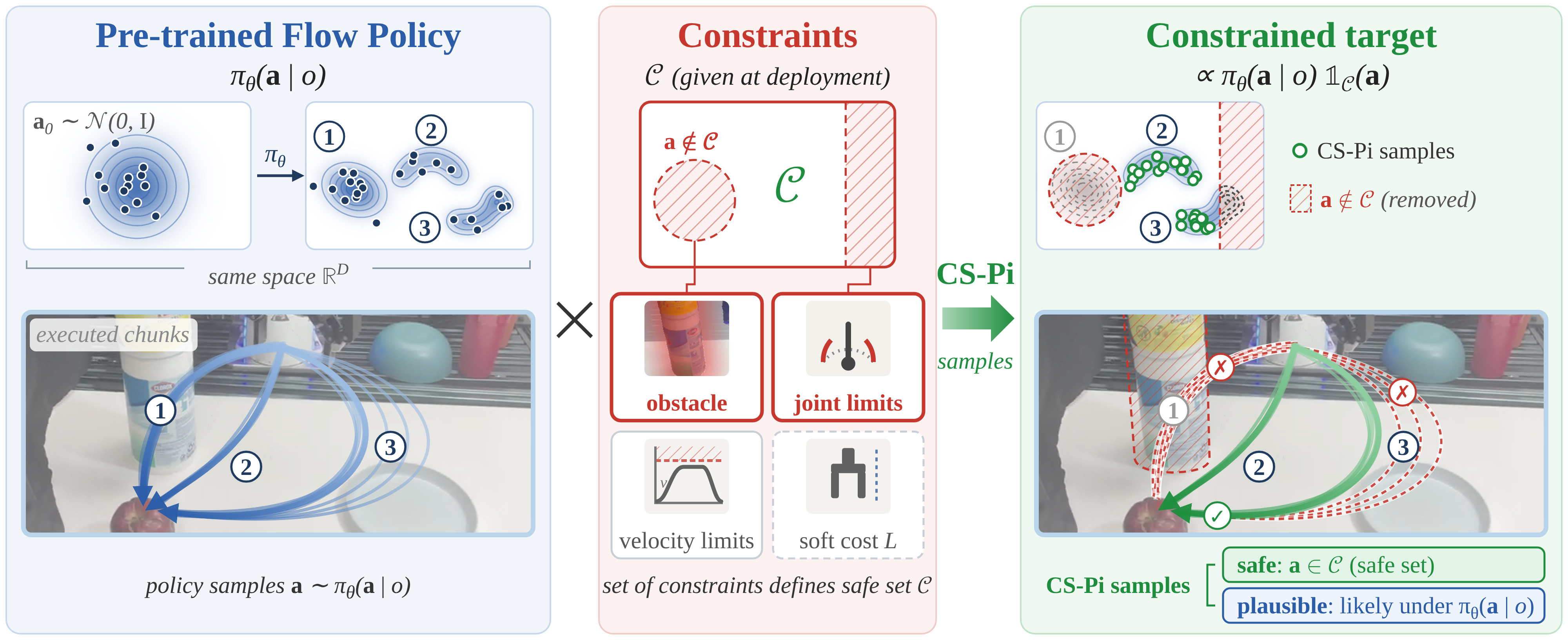}
\caption{\textbf{Sampling a flow-matching policy under a deployment-time constraint.}
\emph{Left:} The policy $\pth$ denoises Gaussian noise
$\mathbf{a}_0\sim\N(\mathbf{0},I)$ ($t{=}0\to1$) into an action chunk
$\mathbf{a}\sim\pth$, a short sequence of robot motions. Its draws are
multimodal: three ways to reach the apple, drawn as paths in the real scene and
as modes of the density $\pth(\mathbf{a})$; the most likely passes
through the obstacle.
\emph{Middle:} Constraints such as obstacle avoidance or velocity and joint
limits, and soft costs $L$, are supplied only at deployment. The obstacle
constraint defines a safe set $\C$, the actions that avoid it.
\emph{Right:} We want actions that are \emph{safe} ($\mathbf{a}\in\C$) and
\emph{plausible} (likely under $\pth$), i.e.\ draws from
$\pth(\mathbf{a})\,\indic{\C}(\mathbf{a})$, where $\indic{\C}$ is $1$ on $\C$
and $0$ elsewhere: the unsafe mode is removed and the others keep their shape.
\AMR{} (ours) targets this distribution without retraining the policy.}
\label{fig:problem}
\end{figure}
\vspace{-2pt}

Among such training-free samplers, the most common is \emph{gradient guidance}~\citep{janner2022planningdiffusionflexiblebehavior,ho2022classifierfreediffusionguidance,song2026omniguideuniversalguidancefields}, which nudges each denoising step along the gradient of a cost on the action. Guidance has two shortcomings. First, safety constraints are typically \emph{hard constraints}: an action either satisfies them or violates them, with no degree in between. An action either touches the wine glass or it does not; this yes-or-no answer says nothing about which way to move the arm, so a hard constraint offers no gradient to follow. Guidance must instead use a smooth cost, such as a collision cost that grows as the arm nears the glass, which discourages contact but does not rule it out. Second, even a smooth cost is meaningful only on a clean action, while guidance acts on noisy intermediate states. Guidance therefore evaluates the cost at a \emph{proxy}: the policy's running estimate of the clean action, an average over every clean action the noisy state could still become. This proxy is reliable only when the cost changes slowly across those actions, so sharpening the cost toward the true constraint makes guidance less reliable, not more. The proxy fails when the policy is uncertain between distinct modes, such as passing left or right of the glass. The average of the two can pass straight through the glass, an action the policy would never sample. The gradient evaluated there may point towards neither mode (App.~\ref{app:guidance-fail}).

We instead build on rejection and projection, two classical mechanisms that act only on the \emph{final} sampled action rather than guiding the intermediate noise levels. Rejection sampling redraws complete actions from the policy until one satisfies the constraint: every output is plausible, but the wait is intractable when safe actions are rare. Projection maps a sampled action to the nearest constraint-satisfying point: it is efficient, but the projected action need not be plausible and can be jerky or task-irrelevant. Our key insight is that flow-matching sampling lets us combine the two, keeping projection's efficiency and rejection's plausibility. When a projected action is implausible, we need not take rejection's drastic step of restarting sampling from pure noise; we can instead resample from a \emph{reduced} noise level around the projection, so the policy returns a safe and plausible action near the projection.

\begin{figure}[t]
\centering
\includegraphics[width=\textwidth]{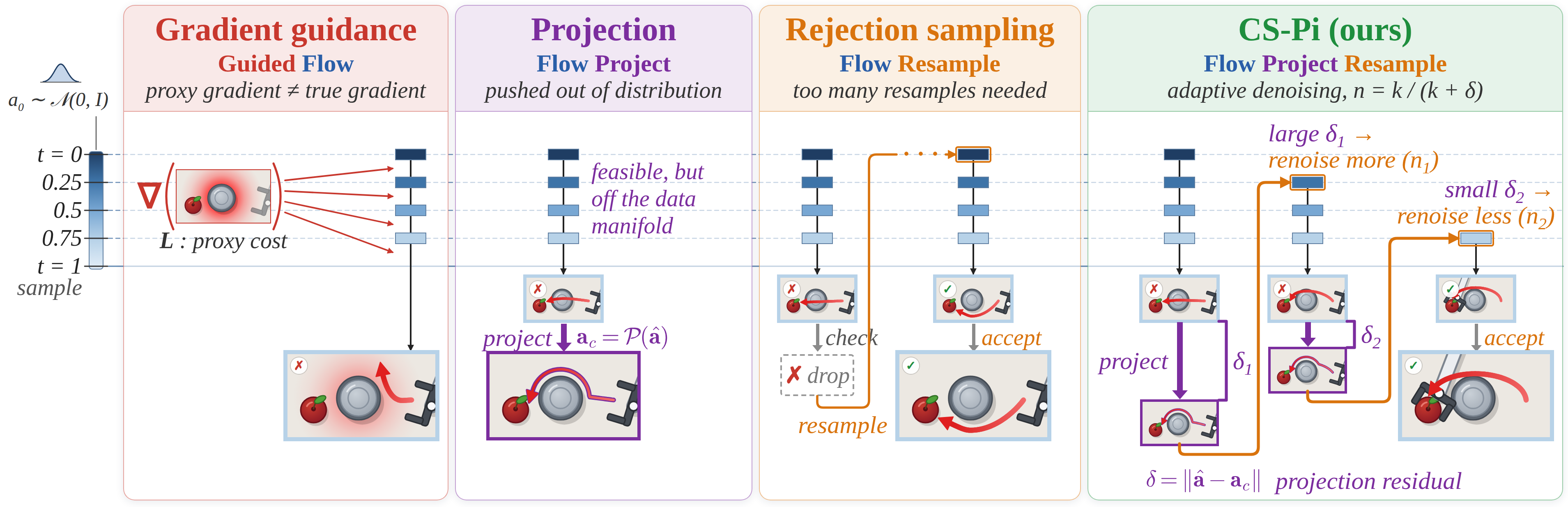}
\caption{\textbf{Sampling a safe action from a flow-matching policy.}
\textbf{Gradient guidance} steers each denoising step with a cost
gradient evaluated at a proxy and can still violate the constraint.
\textbf{Projection} moves the draw to the safe set, yielding an action
the policy would rarely sample. \textbf{Rejection sampling} redraws from
pure noise until a draw is safe. \textbf{\AMR{}} projects the draw,
renoises the projection to time $n=k/(k+\delta)$ set by the projection
residual $\delta$ (larger $\delta$ renoises more), and re-runs the policy
until a draw is safe and plausible.}
\label{fig:concept}
\end{figure}

We operationalize this insight in \AMR{}, a draw-project-renoise sampler (Fig.~\ref{fig:concept}). \AMR{} starts from a single draw from the policy and refines it in a loop. Each iteration projects the current draw onto the safe set and, rather than executing the possibly implausible projection, adds noise back to it, scaled to how far the projection moved the draw. It then re-runs the unmodified policy from that noise level to obtain the next draw. The loop ends once a draw is safe. Each draw comes from the policy and tends to land closer to the safe set than the last, so the draws approach the safe set while remaining plausible, at a compute cost far below rejection and with no constraint gradient at any noisy state.
We first present \AMR{} for hard constraints, then extend it to continuous costs, such as an orientation preference, by replacing only the projection step. The same change lets \AMR{} combine hard constraints with costs. Across a 2D diagnostic, a 7-DoF shelf pick-and-place with deployment-time obstacles, the orientation preference above, and a zero-shot Franka deployment, \AMR{} is the only sampler that satisfies the constraint while staying plausible wherever the constraint permits; on the shelf it reaches $81\%$ safe success (completing the task without collision) against $55\%$ for projection and $32\%$ for gradient guidance.

